\documentclass[10pt,twocolumn,letterpaper]{article}

\usepackage{cvpr}              

\newcommand\nonumfootnote[1]{%
\begingroup%
    \renewcommand\thefootnote{}\footnote{\hspace{-4pt}#1}%
    \addtocounter{footnote}{-1}%
\endgroup%
}

\definecolor{cvprblue}{rgb}{0.21,0.49,0.74}
\usepackage[pagebackref,breaklinks,colorlinks,allcolors=cvprblue]{hyperref}

\def\paperID{2897} 
\def\confName{CVPR}
\def\confYear{2025}

\title{BACON: Improving Clarity of Image Captions via Bag-of-Concept Graphs}

\author{Zhantao Yang$^{1,2}$$^\star$, Ruili Feng$^{2}$$^{\star\diamond}$, Keyu Yan$^{2}$, Huangji Wang$^{1}$, Zhicai Wang$^{2}$ \\
    Shangwen Zhu$^{1}$, Han Zhang$^{1,2}$, Jie Xiao$^{2}$, Pingyu Wu$^{2}$, Kai Zhu$^{2}$, Jixuan Chen$^{2}$ \\
    Chen-Wei Xie$^{2}$, Yue Yang$^{3}$, Hongyang Zhang$^{4}$, Yu Liu$^{2}$, Fan Cheng$^{1}$$^\dagger$ \\ $^1$Shanghai Jiao Tong University, $^2$Alibaba group \\$^3$University of Pennsylvania, $^4$University of Waterloo \\
    \footnotesize\texttt{\{ztyang196, ruilifengustc, yankeyu66, zhushangwen6, hzhang9617, jiexiao916, wpy364755620\}@gmail.com}
    \\
    \footnotesize\texttt{zhicaiw@outlook.com}\quad
    \footnotesize\texttt{\{kaizhustc, chenjixuan.cjx, eniac.xcw, ly103369\}@alibaba-inc.com}\\
    \footnotesize\texttt{yueyang1@seas.upenn.edu}\quad \footnotesize\texttt{hongyang.zhang@uwaterloo.ca}\quad \footnotesize\texttt{chengfan85@gmail.com}\\
    \href{https://ztyang23.github.io/bacon-page}{https://ztyang23.github.io/bacon-page}}
    
\usepackage{overpic, multirow, amsmath, booktabs}

\newcommand{\captioner}{\textsc{LLaVA(BACON)-Captioner}\xspace}
\newcommand{\methodabbr}{\textsc{BACON}\xspace}

\newcommand{\Method}{Bag-of-Concept Graph\xspace}
\newcommand{\dset}{\textsc{ECO}\xspace}
\newcommand{\tocite}[1]{\textcolor{red}{[TO CITE]}}

\usepackage[accsupp]{axessibility}  

\begin{document}

\maketitle
\begin{abstract}

Advancements in large Vision-Language Models have brought precise, accurate image captioning, vital for advancing multi-modal image understanding and processing. Yet these captions often carry lengthy, intertwined contexts that are difficult to parse and frequently overlook essential cues, posing a great barrier for models like GroundingDINO and SDXL, which lack the strong text encoding and syntax analysis needed to fully leverage dense captions.
To address this, we propose \methodabbr, a prompting method that breaks down VLM-generated captions into disentangled, structured elements such as objects, relationships, styles, and themes. This approach not only minimizes confusion from handling complex contexts but also allows for efficient transfer into a JSON dictionary, enabling models without linguistic processing capabilities to easily access key information.
We annotated 100,000 image-caption pairs using \methodabbr with GPT-4V and trained an LLaVA captioner on this dataset, enabling it to produce \methodabbr-style captions without relying on costly GPT-4V. Evaluations of overall quality, precision, and recall—as well as user studies—demonstrate that the resulting caption model consistently outperforms other SOTA VLM models in generating high-quality captions.
Besides, we show that \methodabbr-style captions exhibit better clarity when applied to various models, enabling them to accomplish previously unattainable tasks or surpass existing SOTA solutions without training. For example, \methodabbr-style captions help GroundingDINO achieve 1.51$\times$ higher recall scores on open-vocabulary object detection tasks compared to leading methods.

\end{abstract}

\nonumfootnote{$\dagger$ Corresponding author, $\star$ Equal contribution, $\diamond$ Project leader}    
\section{Introduction}

Image captioning is a crucial task that plays a significant role in fields such as text-to-image generation and multi-modal understanding. The rapid advancement of vision-language models (VLMs)~\citep{liu2023visual,bai2023qwen,chen2023internvl,gpt4v,podell2023sdxl,betker2023improving} has empowered these models to serve as highly effective captioners, producing accurate and detailed descriptions. 
However, their generated captions often suffer from issues such as information entanglement and excessive length, collectively referred to as a \textbf{“lack of clarity.”}
This poses challenges for applications that utilize these captions, particularly in models lacking strong text encoders based on large language models (LLMs), such as SDXL~\citep{podell2023sdxl} and GroundingDINO~\citep{liu2023grounding}. 
Both struggle due to their lack of strong text encoders and syntax analysis capabilities; SDXL has difficulty extracting key objects and their descriptions, as well as comprehending lengthy texts, while GroundingDINO has difficulty identifying essential nouns from long texts.
Besides, these captions frequently overlook useful details, such as image style and relationships between objects. 


\begin{figure*}[t]
\centering
\includegraphics[width=\textwidth]{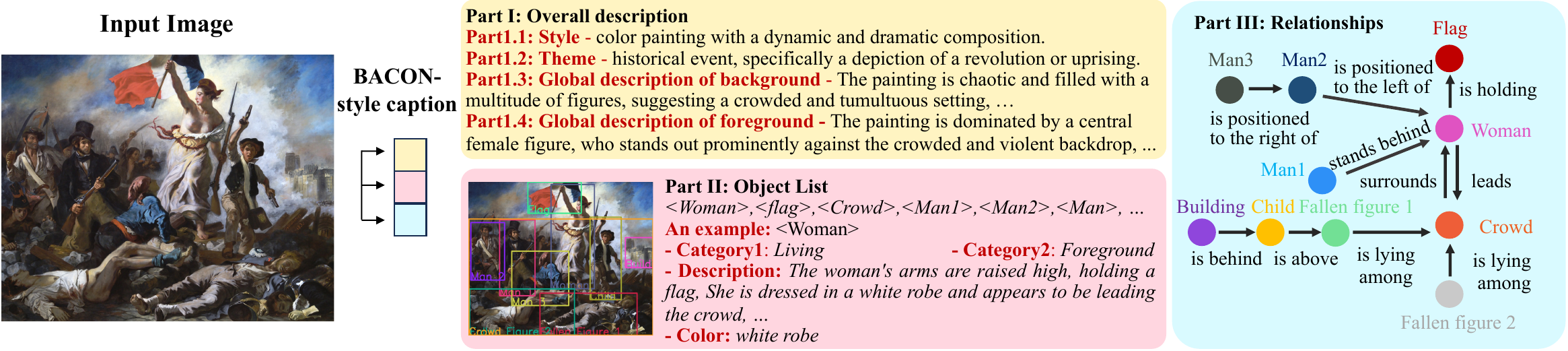}
\vspace{-15pt}
\caption{
\textbf{The \methodabbr-style captions} consist of three components: an overall description, an object list, and relationships. Each object in the object list is accompanied by its category information, detailed description, and color information.
}
\vspace{-10pt}
\label{fig:bacon_structure}
\end{figure*}

To address this challenge, this paper proposes \methodabbr—a neat and efficient approach for extracting \textbf{element-wise} and \textbf{structured} captions from VLMs, resulting in clearer and more complete image captions. The element-wise method breaks down images into basic elements, facilitating information disentanglement and enhancing interpretability. Its structured format allows for quick access to individual elements, alleviating issues related to syntax and order. Additionally, by incorporating all important elements, \methodabbr typically produces more complete captions.
As illustrated in \cref{fig:bacon_structure}, a \methodabbr-style caption composes of three key parts: 1)  an \textbf{overall description} capturing the image’s style, theme, and key features; 2) a detailed \textbf{object list} with labels and descriptions for each item; 3) the \textbf{relationships} between these objects. 

In practice, to implement \methodabbr on VLMs and generate such captions, we first transform the \methodabbr-style captions into a VLMs-readable string format by concatenating the basic elements in a fixed order and using special symbols to divide them. We then apply In-Context Learning (ICL) techniques~\citep{brown2020language} to guide VLMs in producing outputs in our structured format.
Using \methodabbr, we create a dataset called \textbf{Enumerate Common Objects in Context} (\dset), which contains over 100k image-annotation pairs. We also fine-tune a LLaVA model~\citep{liu2023visual} on \dset as the specialized captioner called \captioner.

Experimental results across various benchmarks, including a newly proposed benchmark called Caption Question Answering (CQA), demonstrate that \captioner outperforms commonly used VLM-based captioners. Specifically, adapted from VQA, CQA employs a fixed QA model that relies on captions generated from images—rather than the images themselves—to answer questions, using answer accuracy to evaluate the performance of the captioners, as detailed in ~\ref{subsec:exp_caption_eval}. Additionally, we show that \methodabbr-style captions provide greater clarity when applied to different models, enabling them to perform previously unattainable tasks or surpass existing state-of-the-art solutions without the need for additional training. This includes enhancing Grounding DINO~\citep{liu2023grounding} for open-vocabulary object detection, boosting LLaVA~\citep{liu2023visual} for region-based QA, improving SDXL~\citep{podell2023sdxl} for image generation, and advancing SAM2~\citep{ravi2024sam} for dense video captions.

\section{Related Works}\label{sec:app-background}

\noindent\textbf{Image Captioning.}
Image captioning is an important research topic in the field of artificial intelligence, playing a crucial role in multimodal understanding and image generation. Traditional image captioning methods~\citep{anderson2018bottom, mao2016generation, kazemzadeh2014referitgame, sharma2018conceptual, vinyals2015show} typically rely on manually annotated datasets, such as MS COCO~\citep{lin2014microsoft} and Flickr30k~\citep{young2014image}, using deep learning techniques to fit the caption datasets. These methods often evaluate performance based on similarity to the dataset with similarity metrics including BLEU~\citep{papineni2002bleu}, METEOR~\citep{denkowski2014meteor}, ROUGE~\citep{lin2004rouge}, CIDEr~\citep{vedantam2015cider}, and SPICE~\citep{anderson2016spice}. However, limited by the quality of manual annotations, traditional image captioning techniques are gradually being replaced by VLMs with the rapid advancements in this area.

\begin{figure}[t]
\centering
\begin{overpic}[width=0.9\linewidth]{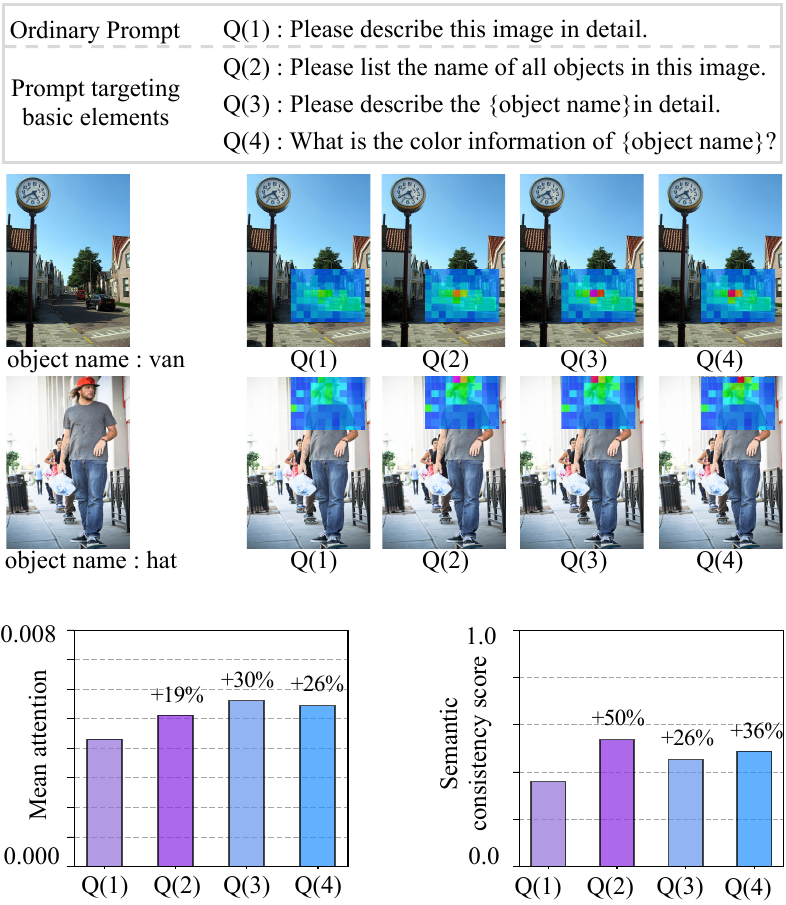}
\put(20,32.5){\small (a) visualization of attention maps}
\put(6,-4){\small (b) mean attention values}
\put(52,-4){\small (c) semantic consistency}
\end{overpic}
\vspace{10pt}
\caption{
(a) Prompts targeting basic elements produce more pronounced attention maps (particularly in crimson) in the target region. (b) Statistical analysis shows the prompts focusing on basic elements have higher average attention values in the target area, indicating an enhanced understanding of VLMs. (c) Prompts targeting basic elements lead to far more consistent answers.
}
\vspace{-10pt}
\label{fig:motivation_exp_1}
\end{figure}

\begin{figure*}[t]
\centering
\begin{overpic}[width=\textwidth]{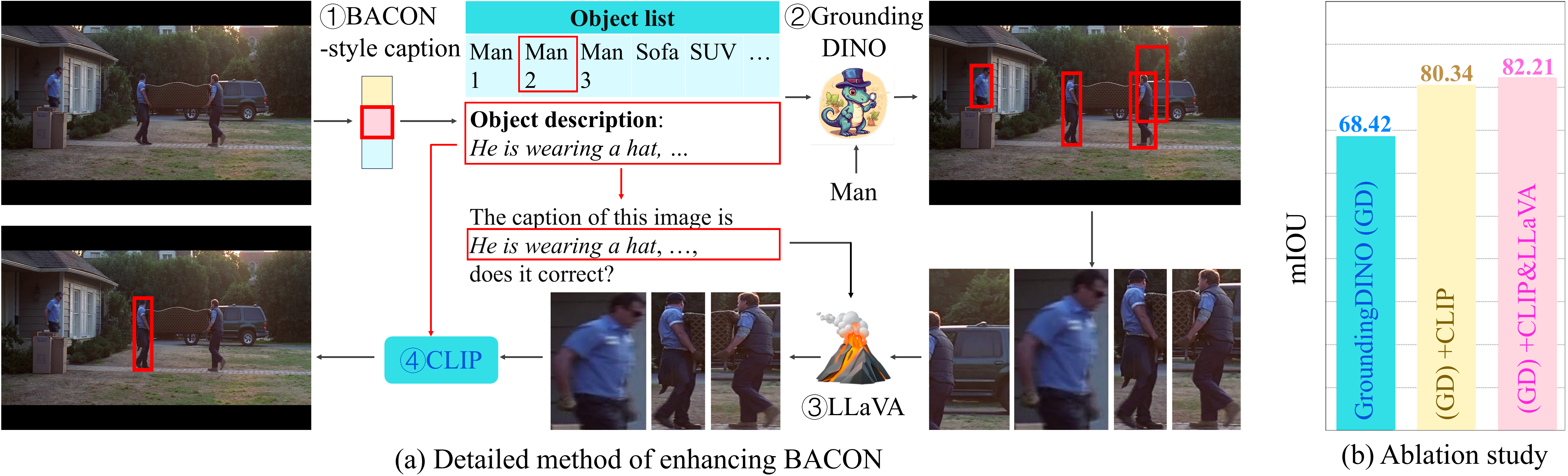}
\end{overpic}
\vspace{-18pt}
\caption{
(a) \textbf{Detailed method} for obtaining bounding boxes for \methodabbr-style captions: 1) Extract \methodabbr-style captions without bounding boxes from images using GPT-4V or \captioner; 2) Generate candidate regions using Grounding DINO given the object name; 3) Use LLaVA to delete clearly incorrect regions; 4) Use CLIP to select the region that best matches the object description. (b) \textbf{Ablation study} conducted on \dset dataset, showing the improvement of introducing CLIP and LLaVA.
}
\label{fig:grounding_dino}
\end{figure*}

\noindent\textbf{Vision Language Models.}
With the rapid development of large language models, vision language models (VLMs)~\cite{gpt4v} have also advanced quickly. These models typically build upon large language models by introducing a vision encoder based on Vision Transformers (ViTs)~\citep{dosovitskiy2020image}. They often train a simple adapter to align the two modalities~\citep{liu2023visual,bai2023qwen,chen2023internvl}. In addition, there are models~\citep{team2023gemini} that do not use adapters and instead directly concatenate visual and textual features, relying on massive datasets and parameter counts to align multiple modalities. The swift progress of VLMs has brought significant innovations to the field of image captioning. Currently, models like GPT-4V excel in image description tasks, significantly outperforming traditional methods based on manually annotated datasets.

\noindent\textbf{Graph-Based Image Captioning Methods.}
Several existing works can generate dense graph-based image captions similar to \methodabbr. However, some lack certain features present in \methodabbr. For example, DOCCI~\citep{onoe2024docci}, DenseFusion~\citep{li2024densefusion}, and ReCap~\citep{li2024llava} cannot provide bounding boxes alongside detailed object captions. Others, like DCI~\citep{urbanek2024picture}, rely on fully manual annotations, making them costly. The closest work to \methodabbr is IIW~\citep{garg2024imageinwords}, which first generates bounding boxes for all objects and then creates descriptions for each. However, since VLMs struggle to recognize and caption small masked images, this approach often requires significantly more annotation effort. In contrast, \methodabbr uses VLMs to extract both object lists and descriptions directly from the entire image before assigning bounding boxes, greatly reducing manual effort. We compare these methods with \methodabbr in \cref{tab:compare_graph_dataset_cameraready}.
\section{\Method}\label{sec:method}

%
In this section, we first discuss the core design principles of \methodabbr from two perspectives to generate ideal image captions: element-wise and structural, as outlined in \cref{subsec:design_bacon}. Next, in \cref{subsec:apply_bacon}, we describe the method for applying \methodabbr to VLMs to generate \methodabbr-style captions, including the method for incorporating grounding capabilities. Finally, in \cref{subsec:bacon_dataset}, we utilize \methodabbr to annotate a new dataset containing over 100k image-annotation pairs, and we fine-tune a LLaVA model on it as a specialized captioner, called \captioner.

\subsection{What We Want from \methodabbr?}\label{subsec:design_bacon}
Previous image captions often suffer from issues such as information entanglement, and excessive length. To address these problems, we propose that an ideal caption should be both element-wise and structured.
\textbf{1) Element-Wise:} This operation aims to disentangle the content of the image into basic elements, which enhances clarity and facilitates a more accurate understanding of the image. Besides, since these basic elements contain only unit information, they are typically much shorter than lengthy image captions and easier to understand.
\textbf{2) Structured:} This characteristic makes it easier to extract of any desired basic elements from the captions, thereby compensating for the limitations in grammatical analysis present in many existing models without LLM-based text encoder.
Furthermore, by considering and incorporating all essential content from the images into basic elements, \methodabbr can prevent potential omissions and yield more complete captions.
Inspired by the truth that the real world can be represented using a scene graph of objects and their relationships~\citep{miller1995wordnet,doddington2004automatic,krishna2017visual,lu2016visual,xu2017scene,johnson2015image,johnson2018image}, we employ a graph structure to design the list of basic elements.

Interestingly, while not our original intention, we find that VLMs may prefer to generate element-wise annotations for two reasons: stronger attention maps of output tokens corresponding to specific image regions (as shown in \cref{fig:motivation_exp_1} (a) and (b)) and higher semantic consistency scores in outputs with repeated requests (as shown in \cref{fig:motivation_exp_1} (c)). The details of numerical experiments (\cref{fig:motivation_exp_1} (b) and (c)) can be found in \cref{subsubsec:app-numerical-motivation}.

\begin{figure*}[t]
\begin{minipage}[c]{0.47\textwidth}
\centering
\vspace{8pt}
\begin{overpic}[width=0.9\textwidth]{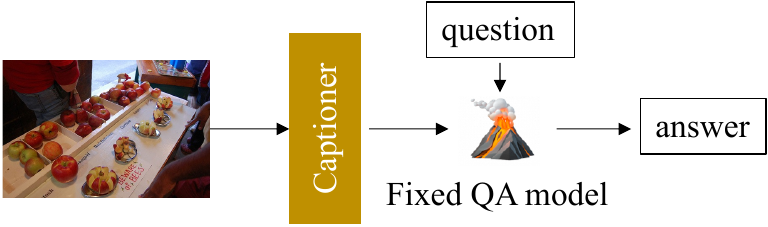}
  \end{overpic}
\centering
\caption{
\textbf{CQA} for evaluating caption quality, where a fixed QA model answers image-related questions based on the caption of the image instead of the image itself.
}
\label{fig:cqa}
\end{minipage}
\hfill
\begin{minipage}[c]{0.478\textwidth}
\centering
\captionof{table}{
\textbf{Results of \captioner (Ours) on CQA benchmarks (illustrated as \cref{fig:cqa})}. The fixed QA model is a pre-trained LLaVA-13B ($\star$ represents that Qwen-14B is used as the fixed QA model)}.
\label{tab:cqa}
\vspace{-12pt}
\footnotesize
\centering
\setlength{\tabcolsep}{2.9pt}
\renewcommand{\arraystretch}{0.7}

\begin{tabular}{l|ccccc}
\toprule
\textbf{Captioner} & \textbf{NLVR2} & \textbf{OK-CQA} & \textbf{CQAv1} & \textbf{CQAv2} & \textbf{NLVR2$^\star$} \\
\midrule
ShareGPT-4V-13B   &  57.5  & 55.4  & 50.7  & 65.4  & 56.0  \\ 
Qwen-VL-max    & 56.8     & 52.1   & 46.0   & 65.6  & 55.0  \\ 
LLaVA-13B    & 56.3  &   54.8 & 50.0  & 64.1  & 55.6  \\ 
\midrule
Ours   &  \textbf{59.1}    & \textbf{56.8}  & \textbf{52.6}  &\textbf{66.4} & \textbf{56.7} \\
\bottomrule
\end{tabular}
\vspace{-.3cm}
\end{minipage}
\vspace{-10pt}
\end{figure*}

\subsection{Method of Implementing \methodabbr.}\label{subsec:apply_bacon}
After introducing the design idea of \methodabbr, we describe the method for implementing it on VLMs like GPT-4V to generate \methodabbr-style captions, as shown in \cref{fig:bacon_structure}. Our approach consists of two steps. First, we convert the \methodabbr-style caption into a VLM-readable string format, where all basic elements are concatenated in a fixed order and separated by special symbols, as illustrated in \cref{fig:string_example}. Second, we employ the ICL~\citep{brown2020language} technique to teach VLMs to output in our specified format. 
We find that a few simplified examples suffice, allowing us to execute the ICL process within a single conversation. In practice, we use GPT-4V as the VLM and provide the final prompt in \cref{fig:instruction}. 

\noindent\textbf{Adding Grounding Capability.}
Due to the sub-optimal grounding capabilities of VLMs, we employ the advanced grounding DINO to obtain the required bounding boxes for objects. 
\methodabbr's structure provides a list of objects required by grounding models, enabling the combination of advanced VLMs for detailing and top-tier grounding models for precise localization within \methodabbr.
Although Grounding DINO provides accurate object positions, names alone fall short of distinguishing objects within the same category. Here, \methodabbr's detailed node descriptions come into play, allowing for precise region identification when used in conjunction with CLIP. Moreover, we improve grounding by using LLaVA to filter out incorrect bounding boxes before the CLIP step. Specifically, we crop candidate regions into individual images and filter them using a CLIP similarity score threshold.
We conducted an ablation study on the \methodabbr benchmark (with details in Section \ref{subsec:bacon_dataset}), and the findings, presented in \cref{fig:grounding_dino} (b), confirm the benefits of incorporating CLIP and LLaVA into our approach. See~\cref{fig:grounding_dino} (a) for an illustration of this process.

\subsection{Collecting the Dataset and Fine-tuning LLaVA.}\label{subsec:bacon_dataset}
Using the above method, we create a dataset called \textbf{Enumerate Common Objects in Context (\dset)} consisting of 103k images and \methodabbr-style caption pairs. The 100k train split is first annotated using GPT-4V and \methodabbr prompt and then comprehensively re-annotated by humans to eliminate ambiguities and inaccuracies. The 3k test split, with 27k objects and 148k relationships, is annotated by separately querying each basic element using VLMs, followed by manual verification to ensure utmost accuracy. See details of collecting \dset in \cref{sec:app-dataset}.

\begin{table}[t]
\caption{
\textbf{Comparison of captioners trained on \dset against those trained on datasets created by other graph-based image captioning methods} reveals that the captioner trained on \dset achieves the best performance, demonstrating that \methodabbr outperforms all competing methods. In these experiments, Qwen-14B is used as the fixed QA model for evaluation.
}
\vspace{-6pt}
\label{tab:compare_graph_dataset_cameraready}
\centering
\setlength{\tabcolsep}{3.1pt}
\renewcommand{\arraystretch}{0.7}
\footnotesize
\begin{tabular}{c|c|c|c|c|c|c}
    \toprule
    Datasets & DOCCI & IIW & DenseFusion & Recap & DCI & \dset(ours) \\
    \midrule
    NLVR2 & 55.81 & 54.37 & 54.19 & 54.22 & 56.16 & \textbf{56.74} \\
    OK-CQA & 46.14 & 44.52 & 45.08 & 43.97 & 44.94 & \textbf{48.49} \\
    MMVP & 57.00 & 58.00 & 56.00 & 57.00 & 58.33 & \textbf{59.67} \\
    \bottomrule
\end{tabular}
\end{table}

\begin{figure}[t]
\centering
\includegraphics[width=0.9\linewidth]{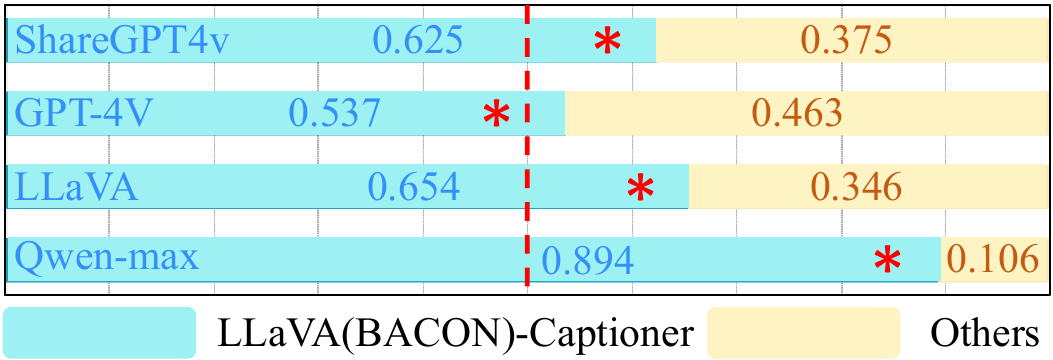}
\vspace{-8pt}
\caption{\textbf{Win rate of pairwise comparisons} between popular VLM-based captioners and \captioner.}
\vspace{-10pt}
\label{fig:user_preference_study}
\end{figure}

We then fine-tune a 13B LLaVA model on the train set as a specialized captioner, called \captioner (see training details in \cref{sec:app-captioner-train}). \captioner performs on par with \methodabbr applied to GPT-4V, having similar distributions of the recognized categories, nouns, and verbs, as shown in \cref{fig:root_analysis}. Furthermore, the manually calculated precision and recall score (as detailed in \cref{subsubsec:exp_user_study}) show \captioner achieve 91\% of precision score and 90\% of recall score of that of \methodabbr applied with GPT-4V. Consequently, \captioner is a viable alternative to GPT-4V for producing \methodabbr-style captions.
Additionally, \captioner can perform useful functions without further fine-tuning, such as interactively modifying the elements of \methodabbr-style captions, transforming ordinary text prompts into \methodabbr-style captions, and planning the locations of all objects.
\section{Experiments}\label{sec:exp}
In this section, we evaluate the image caption produced by applying \methodabbr by first directly measuring the caption quality and then incorporating the captions to various models and enabling them to accomplish previously unattainable tasks or surpass existing state-of-the-art solutions \textbf{without training}. Specifically, we propose a new benchmark adapted from Vision Question Answering (VQA) to directly evaluate the performance of image caption leveraging the advances of Large Language Models, which we introduce in detail in \cref{subsec:exp_caption_eval}. Considering the limiting and space, implementation details and introduction to some experiment settings are placed in \cref{sec:app-experiments}.

\begin{table}[t]
\caption{
\textbf{Manually calculated precision \& recall scores} of other VLM-based captioners and \captioner.}
\vspace{-6pt}
\label{tab:user_study_text}
\centering
\setlength{\tabcolsep}{6pt}
\renewcommand{\arraystretch}{0.7}
\footnotesize
\begin{tabular}{lcc}
\toprule
\textbf{Method} & \textbf{Precision} & \textbf{Recall} \\
\midrule
LLaVA     & 36.4$\pm$1.5$\%$   & 59.2$\pm$4.7$\%$  \\
ShareGPT-4V     & 23.2$\pm$3.8$\%$   & 55.3$\pm$2.1$\%$  \\
Qwen-VL-max    & 35.2$\pm$5.9$\%$     & 57.5$\pm$2.0$\%$   \\
GTP-4V    & 21.5$\pm$0.7$\%$     & 70.6$\pm$ 13.4$\%$  \\
\midrule
\captioner   & $56.2\pm 4.2\%$     & $82.8\pm 8.3\%$  \\
\methodabbr + GPT-4V   & $\bf61.9\pm 3.6\%$     & $\bf92.4\pm 6.8\%$  \\
\bottomrule
\end{tabular}
\vspace{-6pt}
\end{table}

\subsection{Evaluating Image Caption Quality}\label{subsec:exp_caption_eval}

This section employs three distinct evaluation methods to assess the quality of captions produced by \methodabbr. We use LLaVA-13B as our default captioner and denote \captioner as its fine-tuned version in \dset. These evaluation tasks include our newly proposed evaluation approach, Caption Question Answering (CQA) in \cref{subsubsec:exp_cqa}, and open-vocabulary scene graph generation framework in \cref{subsub:exp_ovsgg}, along with comprehensive user studies in \cref{subsubsec:exp_user_study}.

\subsubsection{Analyzing Quality by Caption Question Answering}\label{subsubsec:exp_cqa}
\noindent\textbf{Caption Question Answering (CQA)} The VQA benchmark is broadly used in evaluating the outputs of VLMs. Yet they are not suitable for evaluating captioner models, as 1) the query in VQA benchmarks is very different from image caption, thus they may hardly indicate the ability of image caption of the captioners; 2) captioner models may sacrifice their general VLM capability to enhance image caption ability. So instead, we remove the image in VQA and replace it with its caption produced by different caption methods. The caption and original query are then sent to a fixed QA model to answer the query purely based on text captions. We call this task the \textbf{Caption Question Answering (CQA)} (refer to the illustration in \cref{fig:cqa}). 
Since the correctness of CQA is solely determined by the quality of captions generated by captioners, CQA can serve as an effective evaluation tool for captioners.

\begin{table}[t]
\caption{
\textbf{Comparison of open-vocabulary object detection} among ours (\captioner + Grounding DINO), Grounding DINO, open-vocabulary object detection models, and grounding caption models on \methodabbr benchmark. We have calculated error bars for models that exhibit randomness.
}
\vspace{-6pt}
\label{tab:detection}
\centering
\setlength{\tabcolsep}{7.5pt}
\renewcommand{\arraystretch}{0.8}
\footnotesize
\begin{tabular}{lccc}
\toprule
Method & AP50($\uparrow$) & Recall($\uparrow$) & mIOU($\uparrow$) \\
\midrule
OV-DQUO        & 4.7     & 10.7   & 66.5   \\
DE-VIT         & 19.3    & 23.8   & 76.8   \\
Grounding DINO    & 33.1$_\pm$2.5    & 20.2$\pm$0.1   & 75.7$\pm$0.1   \\
\midrule
Next-Chat     & 29.1$\pm$0.1     & 7.7$\pm$0.1   & 67.1$\pm$0.0    \\
Kosmos-2      & 34.2$\pm$4.8    & 13.3$\pm$2.4   & 76.1$\pm$0.4   \\
GLaMM         & 34.3     & 19.8   & 79.6   \\
\midrule
Ours  & $\bf37.7\pm0.9$     & $\bf35.9\pm0.7$   & $\bf79.9\pm0.1$  \\
\bottomrule
\end{tabular}

\end{table}

\begin{table}[t]
\caption{
\textbf{Accuracy in depicting objects ($A_o$) and relationships ($A_r$) in images generated from text prompts}, as evaluated by human. We compare SDXL enhanced by \captioner (ours) with SDXL and DALL-E 3.
}
\vspace{-6pt}
\label{tab:user_study_image}
\centering
\setlength{\tabcolsep}{10pt}
\renewcommand{\arraystretch}{0.8}
\footnotesize

\begin{tabular}{l|ccc}
\toprule
\textbf{Method} & SDXL & DALL-E 3 & ours \\
\midrule
\textbf{$R_o$}($\uparrow$) & 59.2$\pm$4.0$\%$  & 90.1$\pm$4.2$\%$  & $\bf95.2\pm1.1\%$  \\
\textbf{$R_r$}($\uparrow$) & 41.5$\pm$3.5$\%$  & 71.6$\pm$3.4$\%$  & $\bf76.7\pm0.9\%$  \\
\bottomrule
\end{tabular}
\vspace{-10pt}
\end{table}

\begin{figure*}[t]
\begin{minipage}[c]{0.476\textwidth}
\centering
\vspace{20pt}
\begin{overpic}[width=0.99\linewidth]{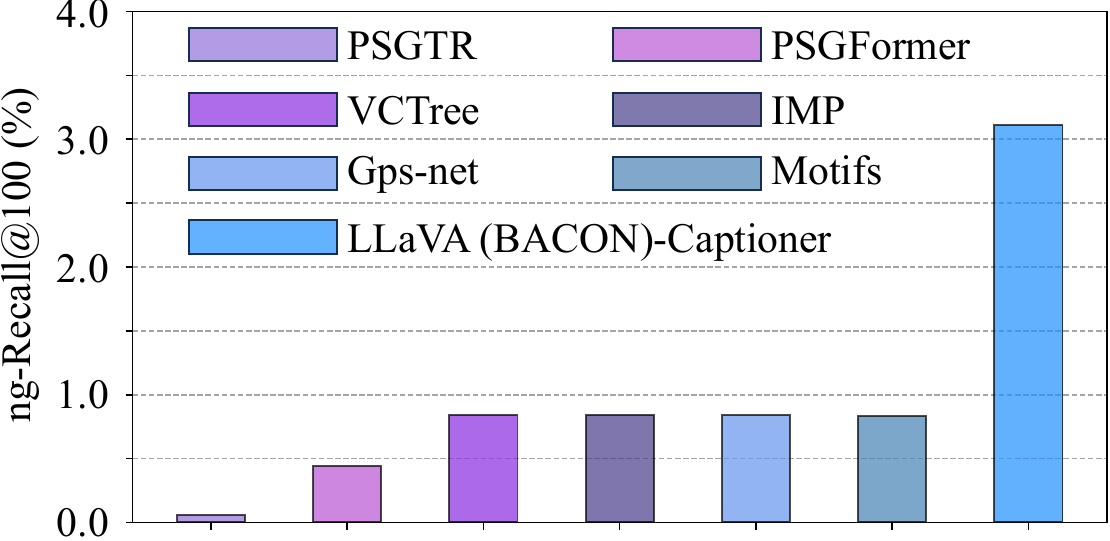}
\end{overpic}
\vspace{-4pt}
\caption{Results of open-vocabulary scene graph generation on VG~\citep{krishna2017visual} benchmarks. \captioner significantly surpasses multiple previous methods.}
\vspace{5pt}
\label{fig:ov_sgg}
\end{minipage}
\hfill
\begin{minipage}[c]{0.47\textwidth}
\centering
\begin{overpic}[width=0.99\linewidth]
{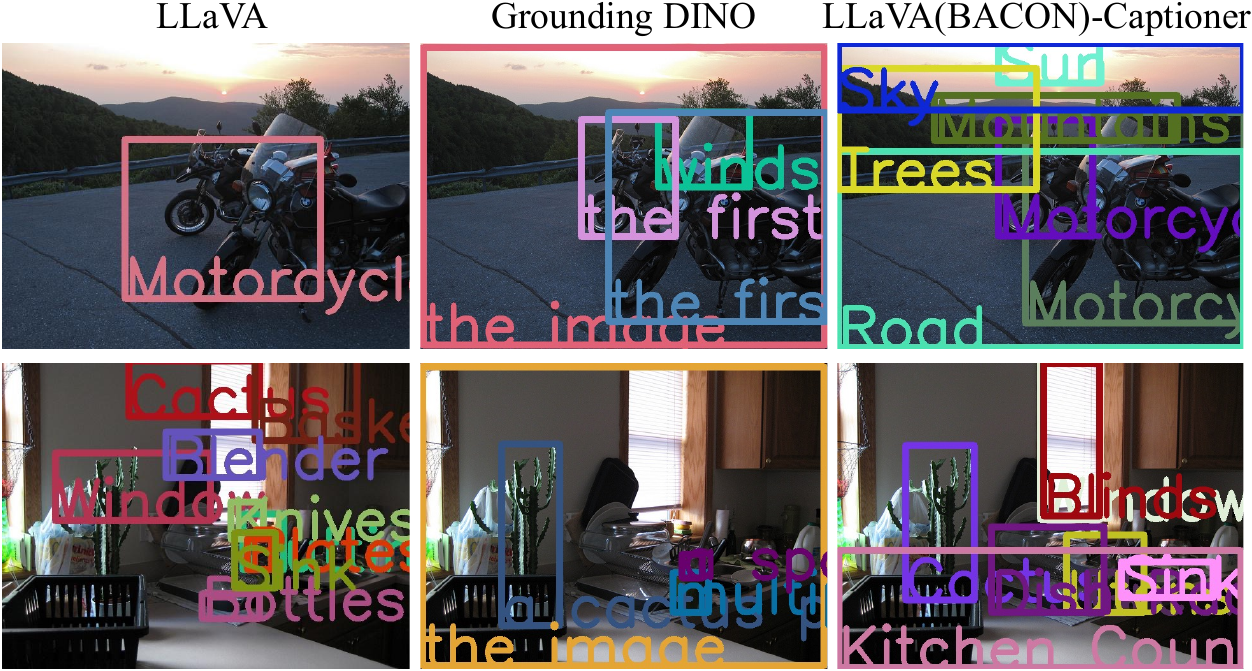}
\end{overpic}
\vspace{-4pt}
\caption{Results of zero-shot open vocabulary object detection. \captioner can significantly improve the performance of grounding DINO.}
\vspace{-10pt}
\label{fig:detection}
\end{minipage}
\vspace{-12pt}
\end{figure*}

We compare our model against three advanced VLM-based captioners, namely LLaVA-13B~\citep{liu2023visual}, Qwen-VL-max~\citep{bai2023qwen}, and ShareGPT-4V~\citep{chen2023sharegpt4v} across four benchmarks: NLVR2~\citep{suhr2018corpus}, VQAv1~\citep{antol2015vqa}, VQAv2~\citep{goyal2017making}, and OK-VQA~\citep{marino2019ok} within the CQA setting. In this paper, we refer to them as CQAv1, CQAv2, and OK-CQA respectively. LLaVA-13B and Qwen-14B~\citep{bai2023qwen} are used as the fixed QA model. The results demonstrated in \cref{tab:cqa} indicate that our approach exhibits significant advantages, underscoring that \captioner generates captions containing much more correct and useful information.

Additionally, we compare the performance of captioners trained on \dset (\captioner) with those trained on datasets created by other graph-based image captioning methods mentioned in \cref{sec:app-background}. In this experiment, we use NLVR2, OK-VQA, and MMVP~\citep{tong2024eyes} as benchmarks. All captioners are fine-tuned from the same pre-trained LLaVA-13B model. As a result, the performance of the fine-tuned captioners reflects both the quality of the datasets and the effectiveness of the captioning methods. The results in \cref{tab:compare_graph_dataset_cameraready} show that \methodabbr outperforms all previous graph-based image captioning methods.

\subsubsection{Analyzing Object \& Relation Accuracy Using Open-Vocabulary Scene Graph Generation}\label{subsub:exp_ovsgg}
Open-vocabulary scene graph generation (OV-SGG) benchmark can be used to measure the accuracy of object and relation recognition. This is because an image can often be explicitly represented using a scene graph composed of objects and their relationships~\citep{krishna2017visual,lu2016visual,xu2017scene,johnson2015image,johnson2018image}. OV-SGG benchmarks aim to measure how many correct (subject-predicate-object) triplets can be extracted from images by the evaluated methods. The details of implementing this evaluation are provided in \cref{sec:app-exp-ovsgg}. We compare the performance of \captioner against several specialized SGG methods, including Motifs~\citep{zellers2018neural}, GPS-Net~\citep{lin2020gps}, VCTree~\citep{tang2019learning}, PSGTR, PSGFormer~\citep{yang2022panoptic}, and IMP~\citep{xu2017scene}, using the widely recognized Visual Genome benchmark~\citep{krishna2017visual}. The results in \cref{fig:ov_sgg} show that \captioner significantly outperforms the previous SGG models, thereby validating the high quality of its output captions.

\subsubsection{Analyzing Precision \& Recall Using User Study}\label{subsubsec:exp_user_study}
We conduct a user preference study to examine the precision and recall score by manual annotation. We compare \captioner and \methodabbr implemented on GPT-4V with LLaVA, Qwen-VL-max, ShareGPT-4V, and GPT-4V, by analyzing captions produced for a randomly sampled set of 200 images from the MSCOCO dataset~\citep{lin2014microsoft}. 
We engage 10 human annotators for manual labeling.
For the precision and recall scores, we first extract all important nouns existing in the captions (see details in \cref{sec:app-exp-user}) and then ask annotators to count the number of objects in the image and the number of correct predictions in the extracted nouns. Then, the precision and recall score can be calculated. 
In the user preference study, annotators select their preferred annotation in pairwise comparisons, ensuring structural aspects are neutralized to prevent any biases. The outcomes, as shown in \cref{fig:user_preference_study,tab:user_study_text}, indicate \methodabbr outperforms all comparisons in general, \textbf{notably predicting more correct objects than multiple popular VLMs even containing GPT-4V}.

\begin{figure*}[t]
\centering
\includegraphics[width=\textwidth]{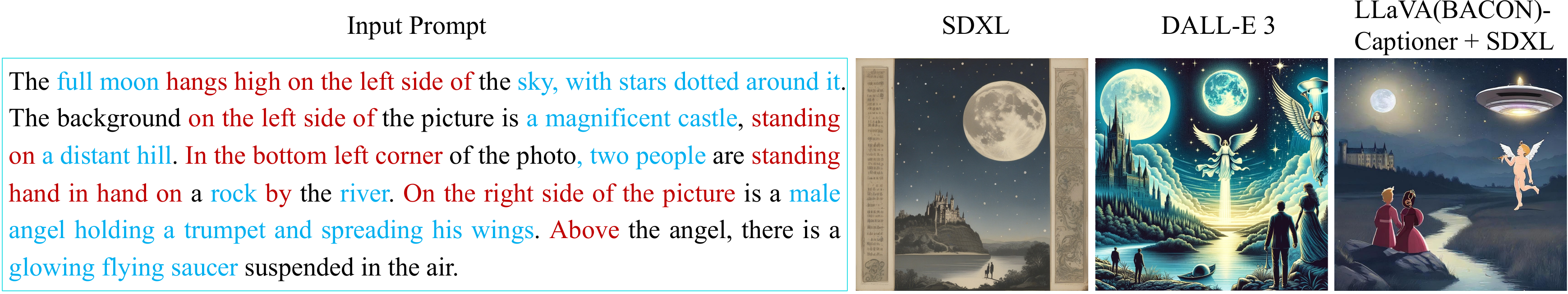}
\vspace{-15pt}
\caption{
\textbf{Comparative examples of image generation} reveal that \captioner enhances advanced generative models like SDXL. SDXL and DALL-E 3 struggle with complex text and fail to produce corresponding images. Remarkably, \captioner not only elevates SDXL's image quality but also markedly boosts its comprehension of intricate instructions, enabling it to surpass DALL-E 3 in terms of accurately generating images aligning with textual directives.
}
\label{fig:image_generation}
\end{figure*}

\subsection{Enhancing Models that Lack LLM-based Text Encoders Without Training}\label{subsec:exp_downstream_exp}

For models that lack a LLM-based text encoder, understanding complex image annotations generated by VLMs becomes a particularly challenging task. This highlights the importance of image captions with better clarity. Fortunately, the adaptable structure of \methodabbr-style captions enhances these models' ability to comprehend complex text and empowers them to perform tasks that were previously beyond their capabilities. In this section, we demonstrate that \methodabbr-style captions provide better clarity by integrating them to assist various models across multiple tasks. These include improving grounding DINO for open-vocabulary object detection (OVD) (\cref{subsubsec:exp_object_detection}), enhancing LLaVA for region-based question answering in a zero-shot setting (ZS-PointQA and ZS-PointingQA) in \cref{subsubsec:exp_pointqa}, boosting SDXL for image generation, detailed in \cref{subsubsec:exp_image_generation}, and enhancing SAM-2\citep{ravi2024sam} for automated multi-object video tracking while seamlessly extending to the task of dense video captioning, as discussed in \cref{subsubsec:exp_video_caption}. It is important to emphasize that all experiments in this section are conducted \textbf{without any additional training.}

\subsubsection{Open-Vocabulary Object Detection}\label{subsubsec:exp_object_detection}
Grounding DINO can process image descriptions to detect mentioned nouns within corresponding images. However, its limited text comprehension often results in the misidentification of irrelevant objects and makes it difficult to distinguish between instances within the same category (as illustrated in \cref{fig:grounding_dino_ovd}). The \methodabbr-style caption mitigates these issues by replacing standard image descriptions with a format that allows Grounding DINO to use item names from the object list as grounding prompts. It can also utilize object descriptions with CLIP~\citep{radford2021learning} to accurately locate target objects, as detailed in \cref{fig:grounding_dino}.

\begin{figure*}[t]
\centering
\includegraphics[width=\textwidth]{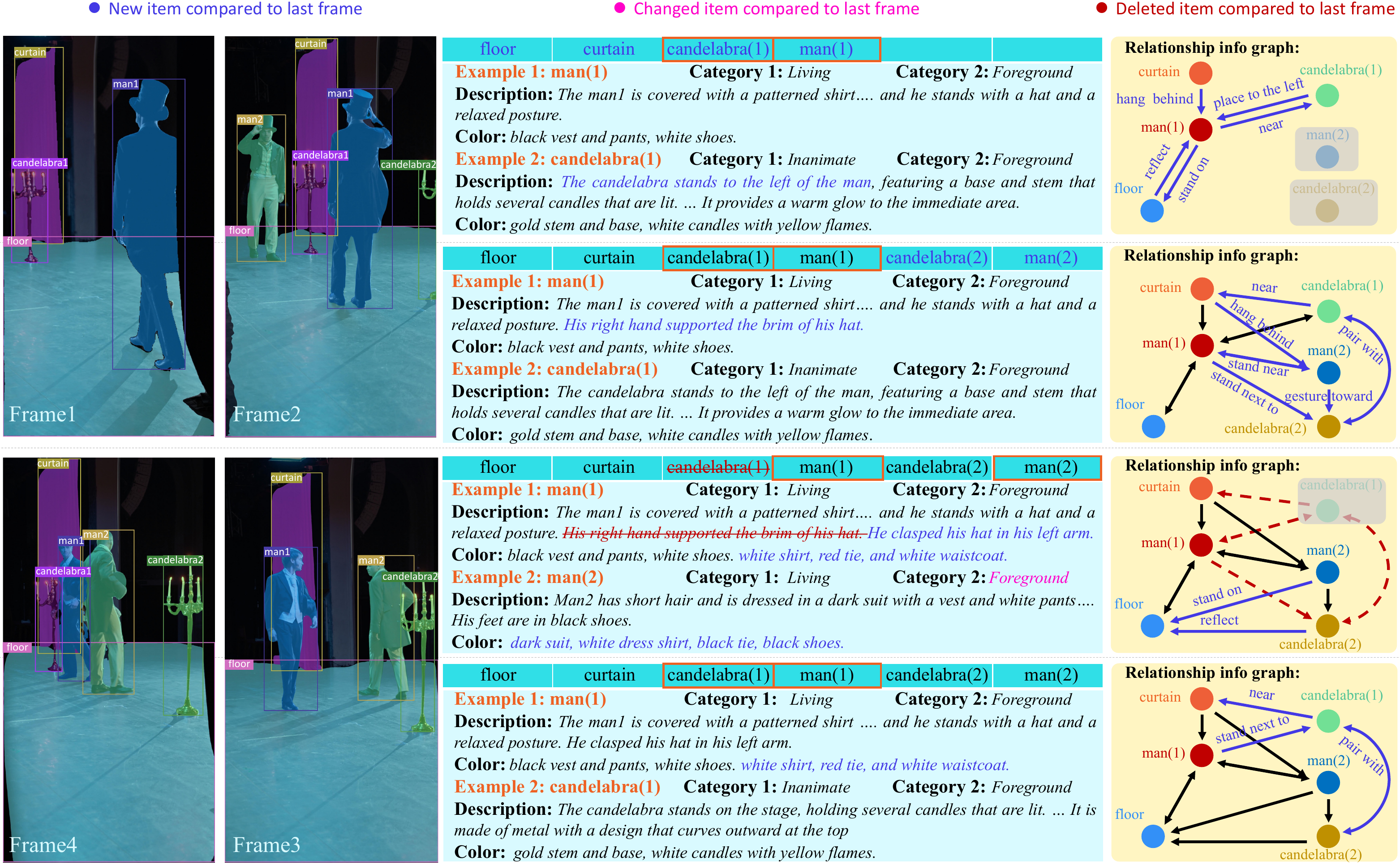}
\caption{
\textbf{An example of \methodabbr on video captioning}, which includes three components: an overall description, an object list, and their relationships, each dynamically evolving over time. With respect to a prior frame, updates are color-coded: new elements in blue, removed in red, altered in pink, and persistent ones in black. \methodabbr thus adeptly captures the temporal changes and salient details of each video frame, while its structured nature potentially aids in downstream model comprehension.
}
\label{fig:video_caption}
\end{figure*}

\noindent\textbf{Evaluation.} Traditional evaluation on OVD task typically categorizes the dataset into base and novel classes, training on base classes and evaluating datasets with novel classes. This resembles a zero-shot rather than an open-vocabulary setting, given the finite number of categories (for example, 80 in COCO). To better reflect an open-vocabulary setting, we evaluate the OVD task on the test benchmark of \dset. Instead of using traditional detection metrics (such as AP50, recall, and mIoU) directly, we modified these algorithms to utilize CLIP similarity between predictions and ground truth for label matching. A successful match is established when the similarity exceeds a predefined threshold without requiring complete correspondence between the prediction and ground truth. In addition to OVD methods including OV-DQUO~\citep{wang2024ov} and DE-VIT~\citep{zhang2024detect}, we also compare the performance with grounding caption models, including GLaMM~\citep{rasheed2023glamm}, Kosmos-2~\citep{peng2023grounding}, Next-Chat~\citep{zhang2023next}.
\noindent\textbf{Results.} 
Results in \cref{tab:detection} show that \methodabbr enhances Grounding DINO to outperform all evaluated methods on the test benchmark of \dset.

\subsubsection{Zero-Shot Region-based Question Answering}\label{subsubsec:exp_pointqa}
\begin{figure}[t]
\centering
\begin{overpic}[width=0.85\linewidth]{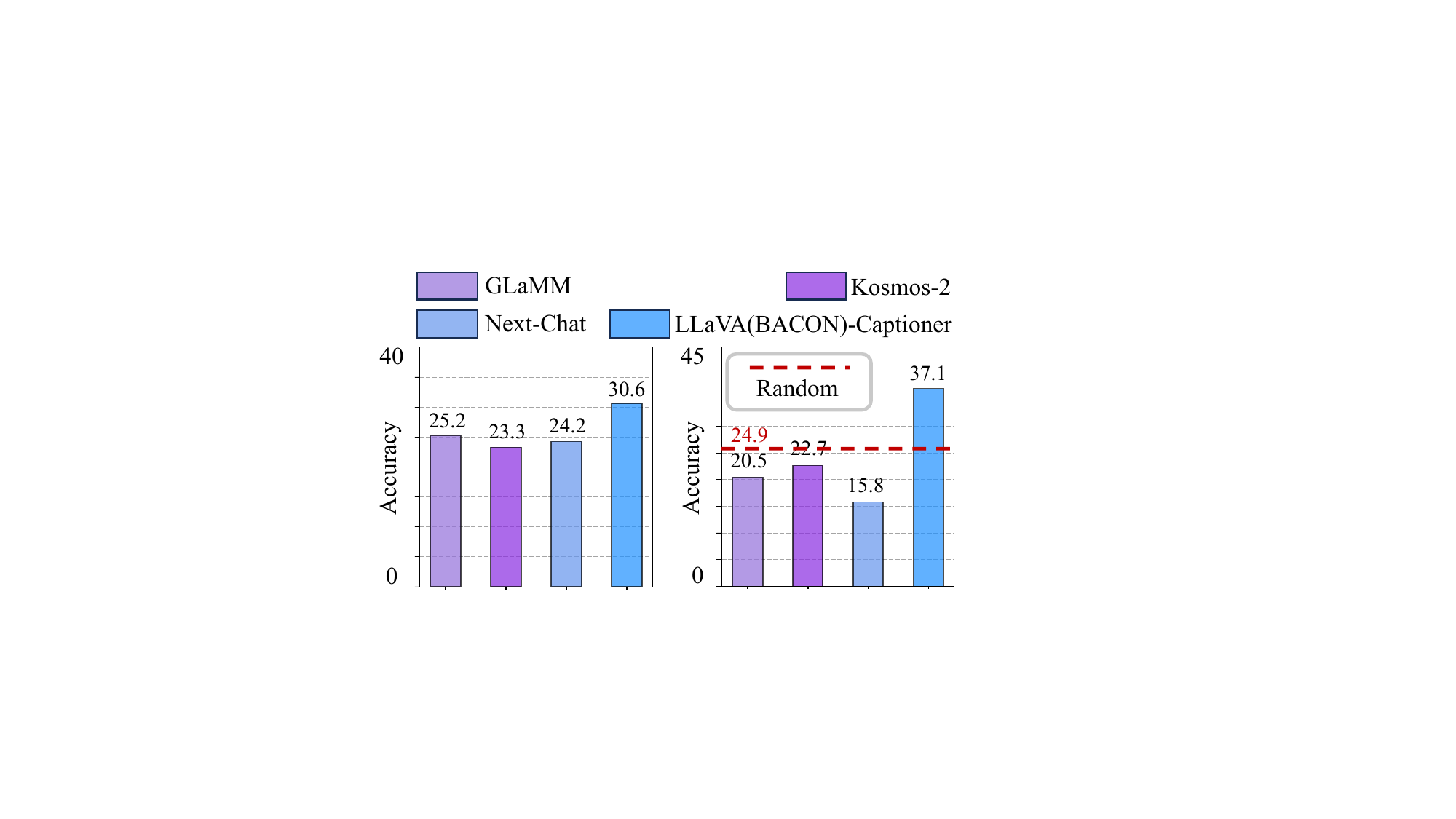}
\put(16,-4){\small (a) PointQA}
\put(67,-4){\small (b) PointingQA}
\vspace{6pt}
\end{overpic}
\caption{
\textbf{Comparison on (a) PointQA and (b) PointingQA} between \captioner and baselines.
}
\vspace{-12pt}
\label{fig:pointing_cqa}
\end{figure}

Zero-shot region-based question answering (QA)~\citep{mani2020point} requires VLMs to answer questions about specific target regions using only the provided image caption, rather than the image itself, as is the case with zero-shot PointQA. Similarly, zero-shot PointingQA~\citep{zhu2016visual7w} tasks VLMs with selecting a relevant region from a set of candidates, again relying solely on the image caption.
These tasks are particularly challenging for VLMs like LLaVA that have not been fine-tuned for such purposes. Fortunately, the \methodabbr-style captions provide a list of objects along with their corresponding regions and detailed descriptions. This information allows us to extract region-relevant descriptions that can effectively address both the zero-shot PointQA and PointingQA tasks. We outline the specific methodology for utilizing \methodabbr-style captions to assist in these tasks in \cref{sec:app-exp-regionqa}.
\noindent\textbf{Evaluation.} We compare our model with grounding caption models, including GLaMM, Kosmos-2, and Next-Chat, that can simultaneously obtain captions and corresponding object bounding boxes. Specifically, we evaluate the zero-shot PointQA task on LookTwice-QA dataset~\citep{mani2020point} (as shown in \cref{fig:pointing_cqa} (a)) and the zero-shot PointingQA task on Visual-7W dataset~\citep{zhu2016visual7w} (as illustrated in\cref{fig:pointing_cqa} (b)). The results demonstrate that \methodabbr-style captions produced by \captioner enable LLaVA to achieve zero-shot region-based QA, a task previously beyond its capabilities while outperforming advanced grounding caption models.

\subsubsection{Multi-Object Video Tracking \& Dense Captioning}\label{subsubsec:exp_video_caption}

The advanced segmentation model SAM-2~\citep{ravi2024sam} supports stable video tracking and holds significant potential for dense video captioning. However, SAM-2 requires text prompts or indicator points to accurately locate target objects, complicating its direct application in video captioning. Fortunately, the \methodabbr-style caption addresses this gap. Specifically, the \methodabbr-style caption includes an object list that specifies the targets for tracking. Additionally, since each object is associated with a mask, the centroids of these masks can serve as effective indicator points. Finally, \methodabbr-style captions provide the essential descriptions needed to accomplish dense video captioning tasks, which exceed the capabilities of SAM-2.
An example illustrated in \cref{fig:video_caption} demonstrates how this approach efficiently captures the continuity and evolution of video content, resulting in coherent and descriptive narration. Additional examples can be found in \cref{fig:app_video_caption_1,fig:app_video_caption_2}.

\subsubsection{Image Generation}\label{subsubsec:exp_image_generation}

Advanced text-to-image generative models like SDXL~\citep{podell2023sdxl} struggle to follow complex text prompts and accurately generate images.
Fortunately, \methodabbr-style caption allows generative models to split the challenge into three easy parts: \textbf{1) Step1.} Given a natural prompt for generation, the trained \captioner can transform it into the \methodabbr-style caption along with a plan of the positions for all objects (discussed in \cref{sec:app-captioner-capability}).
\textbf{2) Step2.} Utilizing the background description part and the detailed description of each object in the object list part, SDXL can generate the background and the important objects separately. \textbf{3) Step3.} Those generated parts are merged according to the planned positions in step1, and then the final image can be refined by common refine methods such as inpainting~\citep{rombach2022high} and SDEdit~\citep{meng2021sdedit}. More details can be found in \cref{sec:app-exp-image-generation}.
\noindent\textbf{Evaluation.}
To quantitatively assess the correlation between the text prompts and the generated images, we conduct a user study involving 10 human annotators and 100 samples. They are required to first annotate the significant objects and relationships mentioned in the text prompts and then count the number of correctly generated ones in the images. Thus we can compute the recall metrics for objects ($R_o$) and relationships ($R_r$). As detailed in \cref{tab:user_study_image}, the results demonstrate that \methodabbr-style caption significantly enhances SDXL's ability to understand and follow complex prompts. Remarkably, it enables SDXL to surpass DALL-E 3 in faithfully reproducing the details specified in the text descriptions. This conclusion can be further supported by the quantitative examples shown in \cref{fig:image_generation} (more instances available in \cref{fig:app_image}).

\section{Conclusion}\label{sec:conclusion}
In this paper, we introduce \methodabbr to significantly enhance the clarity of image captioning. This approach enables models that lack LLM-based text encoders to comprehend complex texts by deconstructing intricate annotations into fundamental elements and organizing them in a graph structure. Utilizing \methodabbr, we construct a dataset at a scale of 100,000 samples. A captioner is then trained on this dataset, showcasing multiple highly valuable capabilities. Extensive experiments demonstrate that our method not only offers significantly improved clarity but also achieves superior quality, as evaluated across various benchmarks, including our newly proposed CQA benchmark.

\section{Limitations}
Despite extensive experiments demonstrating the advantages of \methodabbr, some limitations remain. First, our data collection process relies heavily on manual annotations for correction, as there is currently no fully automated method available to ensure reliability. Second, due to resource constraints, we did not utilize more complex network architectures or larger datasets, resulting in insufficient grounding ability of \captioner. Thus, it remains necessary to use GroundingDINO to obtain accurate bounding boxes.

\nonumfootnote{This work was supported in part by the National Key R\&D Program of China under Grant 2022YFA1005000, in part by the NSFC under Grant 61701304.}

\clearpage
{
    \small
    \bibliographystyle{ieeenat_fullname}
    \bibliography{main}
}

\clearpage
\appendix
\newcommand{\AppendixPrefix}{A}
\renewcommand{\thefigure}{\AppendixPrefix\arabic{figure}}
\setcounter{figure}{0}
\renewcommand{\thetable}{\AppendixPrefix\arabic{table}} 
\setcounter{table}{0}
\renewcommand{\theequation}{\AppendixPrefix\arabic{equation}} 
\setcounter{equation}{0}
\section*{Appendix}

\section*{Overview}\label{sec:app-example}

The supplementary materials consist of four sections:
\begin{itemize}
    \item \textbf{The first section is the details of \methodabbr} (see \cref{sec:app-bacon}), which provides additional details about the implementation of \methodabbr, accompanied by a complete example. The implementation of \methodabbr comprises three parts:
    \begin{itemize}
        \item We introduce the VLM-readable string format mentioned in \cref{subsec:apply_bacon} (see \cref{subsubsec:app-string-format}).
        \item We introduce the specific approach to applying in-context learning (ICL) (see \cref{subsubsec:app-icl})
        \item We discuss how \methodabbr-style captions facilitate integration with Grounding DINO to add grounding capability (see \cref{subsubsec:app-groundingdino}).
        \item We provide detailed numerical experiments to show why VLM prefers element-wise outputs (see \cref{subsubsec:app-numerical-motivation})
    \end{itemize}

    \item \textbf{The second section covers the details of dataset construction} (see \cref{sec:app-dataset}).

    \item \textbf{The third section focuses on \captioner} (see \cref{sec:app-captioner}), including three parts:
    \begin{itemize}
        \item We provide the training details of \captioner (see \cref{sec:app-captioner-train})
        \item We compare the output distribution of \captioner and GPT-4V, demonstrating that \captioner can effectively replace GPT-4V to generate \methodabbr-style captions (see \cref{sec:app-captioner-comparison})
        \item We highlight the interesting capabilities of \captioner beyond generating \methodabbr-style captions, such as interactively editing \methodabbr-style captions, converting ordinary captions into \methodabbr-style, and arranging bounding boxes for all objects.(see \cref{sec:app-captioner-capability})
    \end{itemize}

    \item \textbf{The final section presents additional experimental details} (see \cref{sec:app-experiments}), including five parts:
    \begin{itemize}
        \item We provide additional details of the evaluation of \captioner on the open-vocabulary scene graph generation benchmark, with descriptions of the dataset and evaluation metrics (see \cref{sec:app-exp-ovsgg})
        \item We provide more details about the user study, which explains how to extract important object nouns from the captions during the user study (see \cref{sec:app-exp-user})
        \item We outline how \methodabbr-style captions assist LLaVA in zero-shot region-based question answering (see \cref{sec:app-exp-regionqa})
        \item We provide methodological specifics for using \methodabbr-style captions with SAM-2 for dense video captioning, along with more examples (see \cref{sec:app-exp-videotrack})
        \item We describe how \methodabbr-style captions help SDXL with image generation, accompanied by additional examples. (see \cref{sec:app-exp-image-generation})
        \item We additionally provide the experiments on \methodabbr-style captions enhancing the multi-modal understanding capabilities of VLMs. (see \cref{sec:app-exp-multimodal-understanding})
    \end{itemize}

\end{itemize}

\begin{figure*}[t]
\centering
\includegraphics[width=\textwidth]{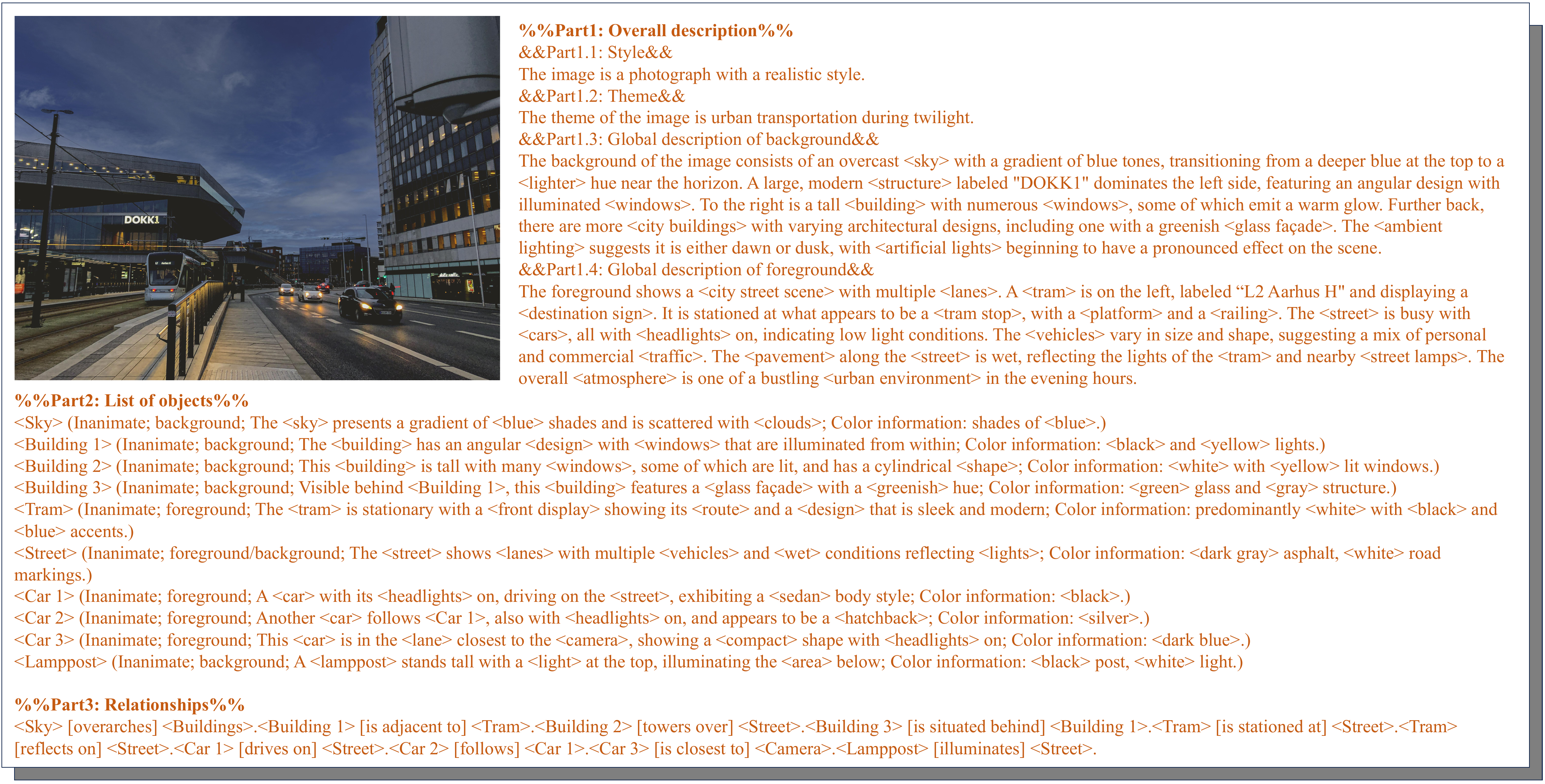}
\caption{
An example of the VLM-readable string format.
}
\label{fig:string_example}
\end{figure*}

\section{Details of \methodabbr}\label{sec:app-bacon}

In this section, we begin by providing comprehensive details about the implementation of \methodabbr in \cref{subsec:app-implement-bacon}. This encompasses a thorough explanation of the VLM-readable string format mentioned in \cref{subsec:apply_bacon}, details on the application of ICL techniques, and additional details on using Grounding DINO to obtain bounding boxes within the \methodabbr-style captions. Lastly, we present a complete example of \methodabbr-style caption that was omitted from the main paper to save space.

\subsection{Detailed method of implementing \methodabbr}\label{subsec:app-implement-bacon}

\subsubsection{The VLM-readable string format}\label{subsubsec:app-string-format}
As discussed in \cref{subsec:apply_bacon}, we convert the \methodabbr-style caption into a VLM-readable string format to enable VLM generation. In this section, we introduce the string format, with an example illustrated in \cref{fig:string_example}. This format is designed by sequentially concatenating all basic elements and separating them with special symbols.
Specifically, we denote main titles with $\%\%$ and subtitles with $\&\&$. When listing objects, we enclose additional details such as category, description, and color in parentheses $()$, with each detail separated by a semicolon ";". The name of an object is marked with $<>$. In describing relationships, we use $<>$ to indicate objects and $[]$ for the predicate. Furthermore, we use $<>$ to highlight important objects within the context, which serves multiple purposes. One of its functions is to post-process the output from GPT-4V, allowing us to remove foreground information from the background description by deleting sentences where the foreground objects appear, or conversely, eliminating background information from the foreground description.
By utilizing these special symbols to separate different sections, we can effortlessly organize the VLM output into a \methodabbr-style caption using regular expressions.

\begin{figure*}[t]
\centering
\begin{overpic}[width=\textwidth]{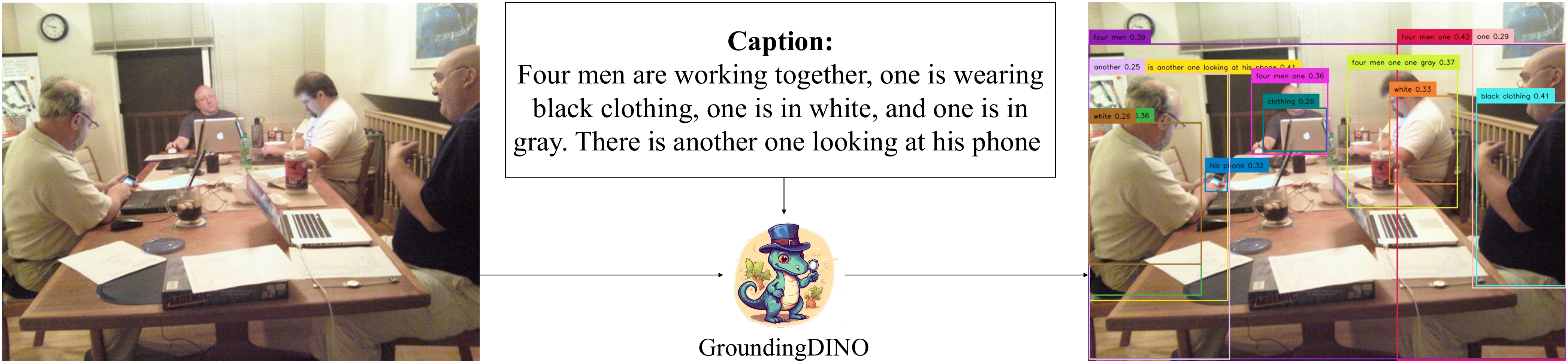}
\end{overpic}
\caption{
\textbf{An example of Grounding DINO performing the object detection task} illustrates its struggles with ambiguous labels and challenges in differentiating between individuals within the same category.
}
\label{fig:grounding_dino_ovd}
\end{figure*}

\subsubsection{The details of implementing ICL technique}\label{subsubsec:app-icl}
\begin{figure*}[t]
\centering
\includegraphics[width=\textwidth]{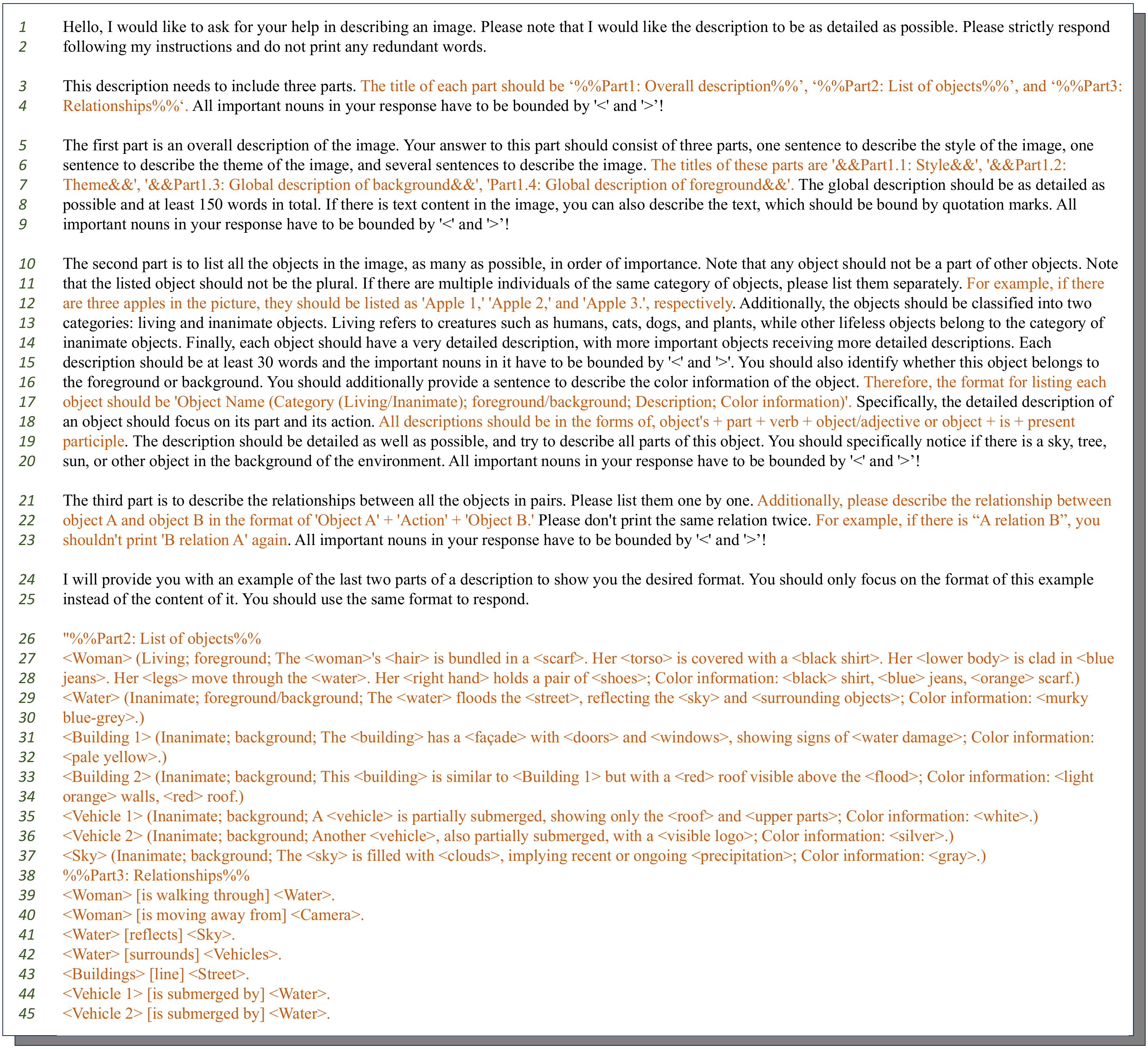}
\caption{
\textbf{The instruction} for GPT-4V to obtain the \methodabbr-style caption from an image. We highlight the parts involving specific examples in orange.
}
\label{fig:instruction}
\end{figure*}

As discussed in \cref{subsec:apply_bacon}, we implemented the ICL technique to guide VLMs in responding with the desired string format outlined in \cref{subsubsec:app-string-format}. In practice, we discovered that GPT-4V does not require an extensive array of examples to understand the required format. Instead, incorporating just a few strategically chosen key examples within the instruction is sufficient. This approach allows us to process ICL within a single conversation, significantly reducing the costs associated with implementing \methodabbr on the expensive GPT-4V. The final instruction is illustrated in \cref{fig:instruction}, where we have highlighted the crucial examples in orange.
Among the provided examples, some are specific while others are more general. We observed that general examples are particularly effective for straightforward structural elements. For instance, brief notations like 'lines 3-4' or 'lines 6-7' sufficiently illustrate the use of special symbols within a section, eliminating the necessity for expansive examples. In lines 16-17, we provide a general example that effectively clarifies the structure of each object, greatly reducing errors made by GPT-4V. To simplify the grasping of object details, we utilized a general example in lines 18-19, which proves adequate for generating simple sentences. Similarly, in lines 21-22, a general example suffices to guide GPT-4V on the fundamental pattern for depicting relationships. Lastly, the general example presented in lines 22-23 helps to prevent GPT-4V from erroneously generating two-way relationship pairs.

However, our stringent requirements concerning content and structure pose significant challenges, even for GPT-4V. As a result, it occasionally makes mistakes, such as omitting special symbols, even when general examples are provided. This highlights the necessity of including specific examples to ensure that GPT-4V accurately comprehends the required structure. For instance, when it comes to numbering items within the same category, we introduced a specific example in lines 11-12. Without this explicit reference, GPT-4V tends to overlook numbering, despite our earlier instructions in lines 10-11.
Additionally, we noticed that while GPT-4V performs well with the format of the first section, it often falters in the second and third parts, complicating the transformation of data into a dictionary format. By offering just one clear example in lines 26-45 for these sections, we significantly increase the likelihood that GPT-4V will generate the correct structure. The implementation of the ICL technique has effectively ensured that nearly all of the 100,000 data entries we’ve collected are formatted correctly and can be seamlessly translated into a dictionary format.

\subsubsection{Why \methodabbr enables the introduction of Grounding DINO to add grounding capability}\label{subsubsec:app-groundingdino}

\begin{figure*}[t]
\centering
\includegraphics[width=0.95\textwidth]{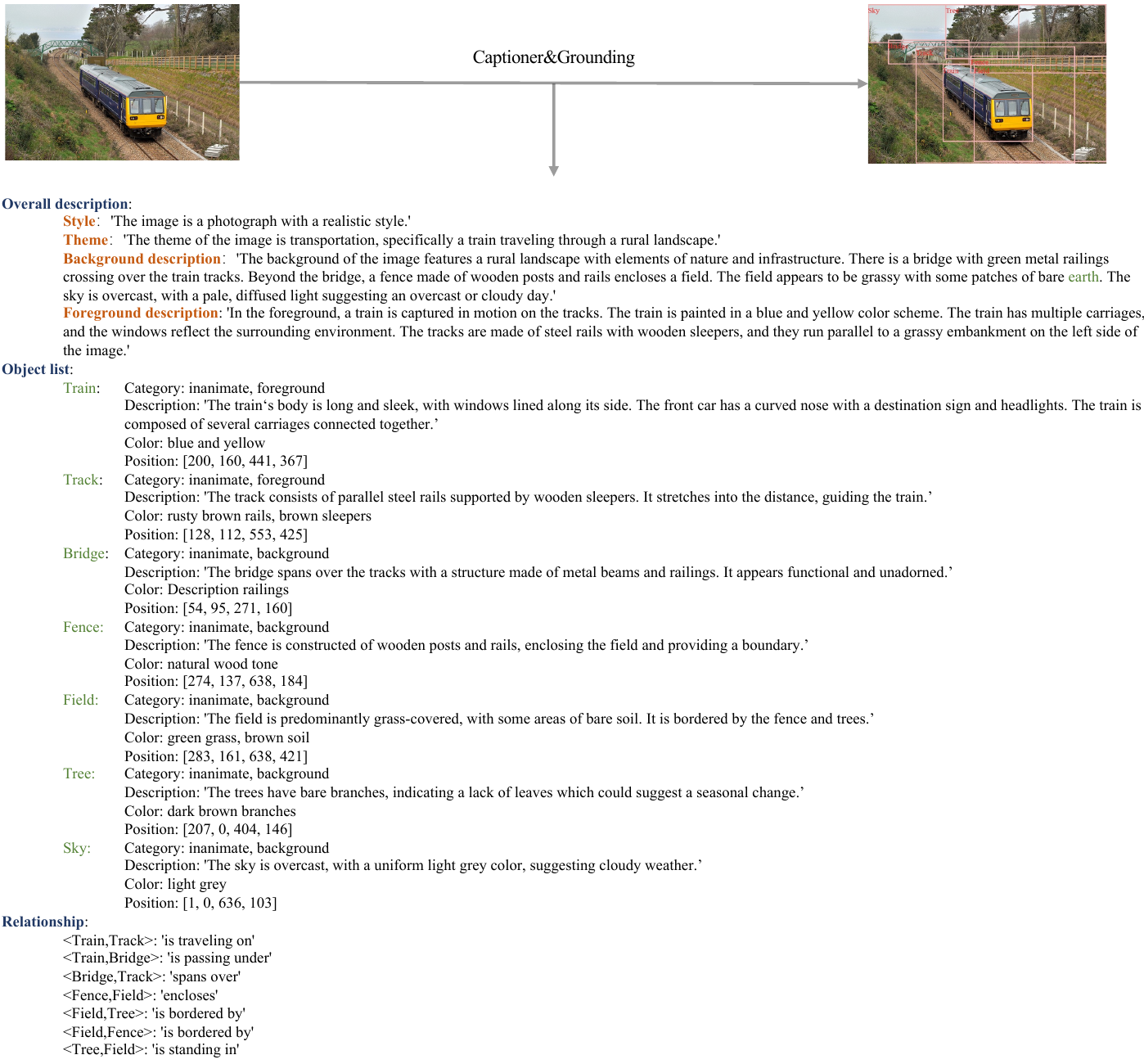}
\vspace{-10pt}
\caption{
\textbf{A complete example of \methodabbr-style caption}.
}
\vspace{-10pt}
\label{fig:app_bacon_example}
\end{figure*}

As discussed in \cref{subsec:apply_bacon} and \cref{subsubsec:exp_object_detection}, Grounding DINO is the leading grounding model, but it faces some issues. 1) First, it lacks the ability to understand long and complex sentences, resulting in its inability to accept more than a single noun as input. When the input is an image description, Grounding DINO struggles to extract nouns, which can lead to bizarre labels. For example, as illustrated in \cref{fig:grounding_dino_ovd}, Grounding DINO produces ambiguous labels such as “one,” "four men one gray," and "another."
2) The second issue, which is more significant, is Grounding DINO's difficulty in differentiating between individuals of the same category. As shown in \cref{fig:grounding_dino_ovd}, while Grounding DINO correctly identifies four people, it remains challenging to determine which individual corresponds to each bounding box, often resulting in vague labels like "four men one." 

Fortunately, the structured format of \methodabbr-style captions enables Grounding DINO to overcome these two challenges. For the first issue, \methodabbr-style captions provide an accurate and comprehensive object list that serves as a reliable source of nouns. Regarding the second issue, as introduced in \cref{fig:grounding_dino}, by leveraging the list of objects provided by \methodabbr-style captions, along with detailed descriptions for each object, it becomes possible to utilize CLIP to make precise distinctions among different individuals sharing the same category label.

\subsubsection{Numerical experiments of why VLMs may prefer element-wise outputs.}\label{subsubsec:app-numerical-motivation}
As discussed in \cref{subsec:design_bacon}, VLMs may prefer element-wise outputs for two reasons: 1) stronger attention maps of output tokens corresponding to specific image regions, and 2) higher semantic consistency scores in outputs with repeated requests. In this section, we present detailed numerical experiments, with the results illustrated in \cref{fig:motivation_exp_1}.

\noindent\textbf{Numerical experiments.} We conduct numerical experiments to show these two key insights. Specifically, we compare a commonly used image captioning prompt, \texttt{"Please describe this image in detail."}, with several questions targeting basic elements, including \texttt{"Please list all objects in this image."}, \texttt{"Please describe the {object name} in detail."}, and \texttt{"What is the color information of {object name}?"}. 
We randomly select 1,000 images from the MSCOCO dataset~\citep{lin2014microsoft} and ask LLaVA these four questions, repeating each one 10 times with different random seeds. We then calculate the mean values of attention maps in the target regions and the semantic consistency scores.

For attention analysis, we extract objects and their corresponding target regions (segmentation masks) from the MSCOCO dataset annotations. For each object, we calculate the mean values of their attention maps within the target region across all output tokens. We then use the maximum value among all tokens as the final score for that object. As shown in \cref{fig:motivation_exp_1} (b), the questions targeting basic elements yield more pronounced attention maps compared to the standard prompt for description, indicating VLMs understand their meaning more firmly.

For the semantic consistency analysis, we compute the similarity evaluated by T5~\citep{raffel2020exploring} of answer sub-sentences among $n$ independent repetitions of question-answering. For each answer $a_i, 1 \le i \le n$, we split it into $k_i$ sub-sentences based on punctuations. We then compute the semantic consistency score between $a_i$ and $a_j$ by $S(a_i, a_j) = \frac{1}{2}(F(a_i|a_j) + F(a_j|a_i))$, where $F(a_i|a_j) = \sum_{m=1}^{k_i} s(a_i^m, a_j)$. Here, $s(a_i^m, a_j)$ is defined as
\def\T5{\mathrm{T5}}
\begin{equation}
s(a_i^m, a_j) =
\begin{cases}
1 & \text{if } \displaystyle\max\limits_{1 \le n \le k_j}\frac{\langle \T5(a_i^m), \T5(a_j^n) \rangle}{\Vert \T5(a_i^m) \Vert_2 \Vert \T5(a_j^n) \Vert_2} \ge \rho \\
0 & \text{else}
\end{cases}
\end{equation}
where $\rho$ is a threshold. $s(a_i^m, a_j)=1$ indicates that a similar sub-sentence of $a_i^m$ can be found in $a_j$. As illustrated in \cref{fig:motivation_exp_1} (c), the three questions targeting basic elements exhibit much higher semantic consistency scores than the normal request for image description, meaning that the VLMs are very confident and clear in what the answers should be.

\subsection{Complete examples of \methodabbr-style captions}\label{subsec:app-bacon-caption}

We provide a complete example of \methodabbr-style caption in \cref{fig:app_bacon_example}

\section{Detailed method of building \dset}\label{sec:app-dataset}

\subsection{Detailed method of building the training set}\label{sec:app-dataset-train}
The structure of \methodabbr-style captions significantly streamlines the workload of annotation during the training data collection process. By breaking down complex descriptions into basic elements, many of which require annotators to simply make a straightforward right-or-wrong judgment, the task becomes highly manageable. For larger segments of information, such as background or foreground descriptions, annotators are instructed to independently assess whether each sentence accurately reflects the image. Additionally, they are asked to identify and add any objects missed by GPT-4V. The design of our structure for object descriptions further aids annotators by simplifying the annotation process; they only need to fill in the corresponding information according to the established format.

\subsection{Detailed method of building the test set}\label{sec:app-dataset-test}

Despite the impressive capabilities of GPT-4V, it may occasionally overlook objects. To achieve maximum accuracy, we employ an entirely different pipeline that involves separately querying each basic element to create the test set for \dset. This method comprises five distinct steps: 1) The Segment anything (SAM)~\citep{kirillov2023segment} model segments all components within the image, helping to prevent the omission of objects. 2) VLMs identify the names of objects in the masked image obtained from the first step. 3) Using the names identified in the second step, VLMs annotate each object in detail. 4) VLMs generate an overall description of the image based on the list of objects derived from the above steps. 5) Images created by randomly pairing two masked images from the first step are fed to the VLMs to identify the relationship between the two objects in the selected masked images.
Annotators are involved in steps 2 through 5. In Step 2, they are responsible for correcting the results returned by the VLMs to accurately identify the object names given the masked images. In Steps 3 and 4, annotators are tasked with verifying the accuracy of each generated sentence. They do not need to add objects, as the Segment Anything (SAM)~\citep{kirillov2023segment} ensures that there are no omissions. In Step 5, annotators must assess whether the identified relationships are correct and add any significant relationships that may have been overlooked by the VLMs.

\begin{figure*}[t]
\centering
\begin{overpic}[width=\textwidth]{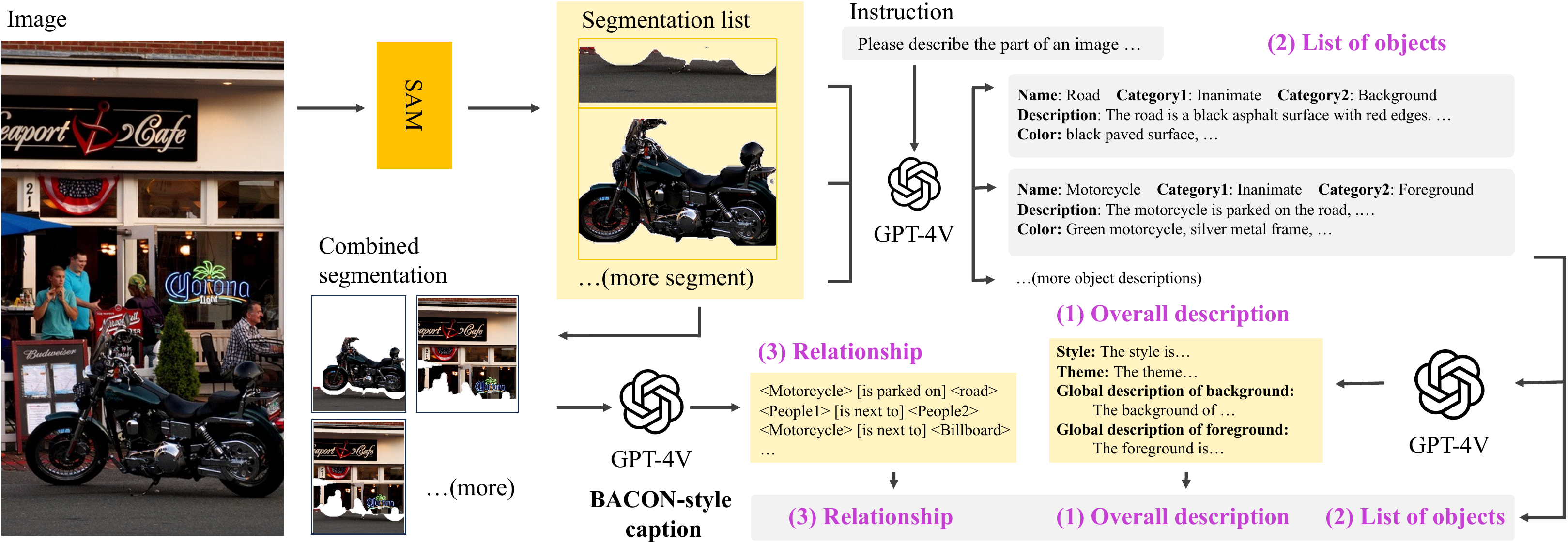}
\end{overpic}
\caption{
\textbf{A detailed overview of the method used to collect the test set of \dset}, segmented into five distinct steps. 1) The SAM model segments all components within the image. 2) VLMs identify the names of objects in the masked image obtained from the first step. 3) Using the names identified in the second step, VLMs annotate each object in detail. 4) VLMs generate an overall description of the image based on the list of objects derived from the above steps. 5) images created by randomly pairing two masked images from the first step are fed to VLMs to identify the relationship between the combined segments. It is important to note that human annotation is required to correct and verify the outputs from steps two through five.
}
\label{fig:test_pipeline}
\end{figure*}

\section{Details of \captioner}\label{sec:app-captioner}

\subsection{Details of training \captioner}\label{sec:app-captioner-train}
\captioner is fine-tuned on \dset using a pre-trained 13B LLaVA model with the Low-Rank Adaptation (LoRA) technique~\citep{hu2021lora}. The total number of LoRA parameters is approximately 0.5 billion. Training is conducted on NVIDIA A100 GPUs and requires about 100 GPU hours, utilizing a learning rate of $2\times 10^{-4}$ for 3 epochs. The batch size used is 16, and the LoRA rank is set to 128. Additionally, we implement a warmup ratio of 0.03 during training.

\begin{table}[t]
\caption{
\textbf{Comparison of plan task} between \captioner (ours) and LayoutGPT~\citep{feng2024layoutgpt} on both MSCOCO and test set of \dset.
}
\vspace{-15pt}
\label{tab:plan}
\begin{center}
\setlength{\tabcolsep}{1.7pt}
\footnotesize
\begin{tabular}{lcccc}
\toprule
\textbf{Dataset} & \textbf{Method} & \textbf{Precision} & \textbf{Recall} & \textbf{mIOU} \\
\midrule
\multirow{2}{*}{MSCOCO}  &LayoutGPT  & 70.1$\%$  & 39.7$\%$ &4.1$\%$\\
& \captioner    & $\bf71.2\%$    & $\bf41.8\%$  &$\bf6.8\%$ \\
\midrule
\multirow{2}{*}{\dset}  &LayoutGPT  & 50.8$\%$  & 29.2$\%$ &9.1$\%$ \\
 & \captioner    & $\bf51.7\%$    & $\bf47.1\%$  &$\bf18.4\%$ \\
\bottomrule
\end{tabular}

\vspace{-15pt}
\end{center}
\end{table}

\subsection{Comparison between \captioner with GPT-4V on obtaining \methodabbr-style captions}\label{sec:app-captioner-comparison}
\begin{figure*}[t]
\centering
\includegraphics[width=\textwidth]{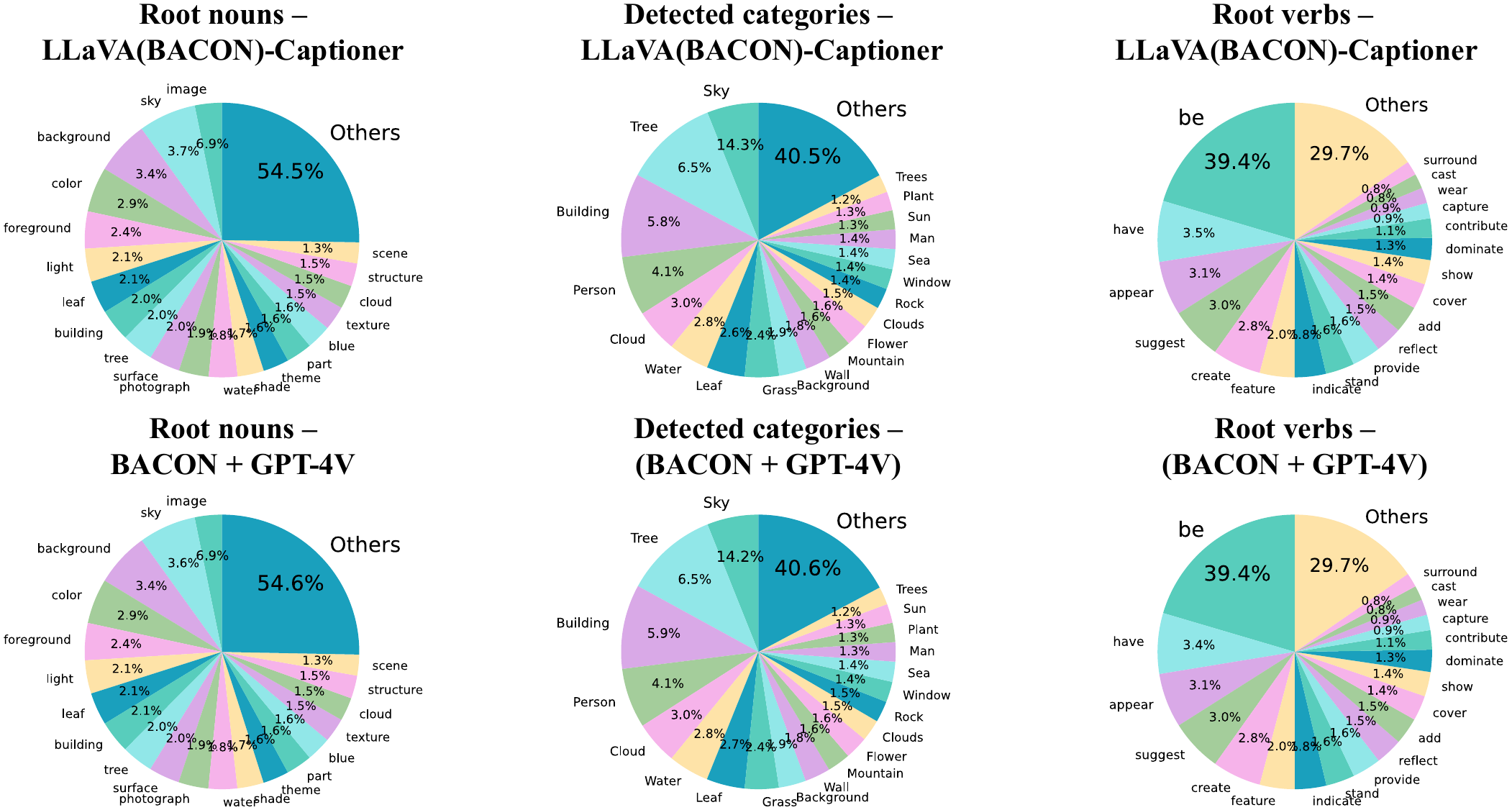}
\caption{\textbf{Comparison of the root words and detected categories generated by \captioner and GPT-4V on the test set of \dset} (certain sections magnified for clearer visualization). The results reveal that the output distribution of \methodabbr closely resembles that of GPT-4V, demonstrating that \captioner can effectively replace the expensive GPT-4V in generating \methodabbr-style captions.
}
\vspace{-10pt}
\label{fig:root_analysis}
\end{figure*}

We present an analysis of the root words and categories identified in the outputs of \captioner, as illustrated in \cref{fig:root_analysis}. The results clearly indicate that the output distributions of \captioner are very similar to those of GPT-4V. Notably, there is a 100\% overlap in the top 100 most frequent nouns, a 99\% overlap for verbs, and a 97\% overlap for categories detected by both GPT-4V and \captioner. This similarity confirms that \textbf{\captioner can effectively take over from GPT-4V in generating \methodabbr} from images, thereby extending our \dset.

\subsection{Additional capabilities of \captioner}\label{sec:app-captioner-capability}

\begin{figure*}[t]
\centering
\begin{overpic}[width=\textwidth]{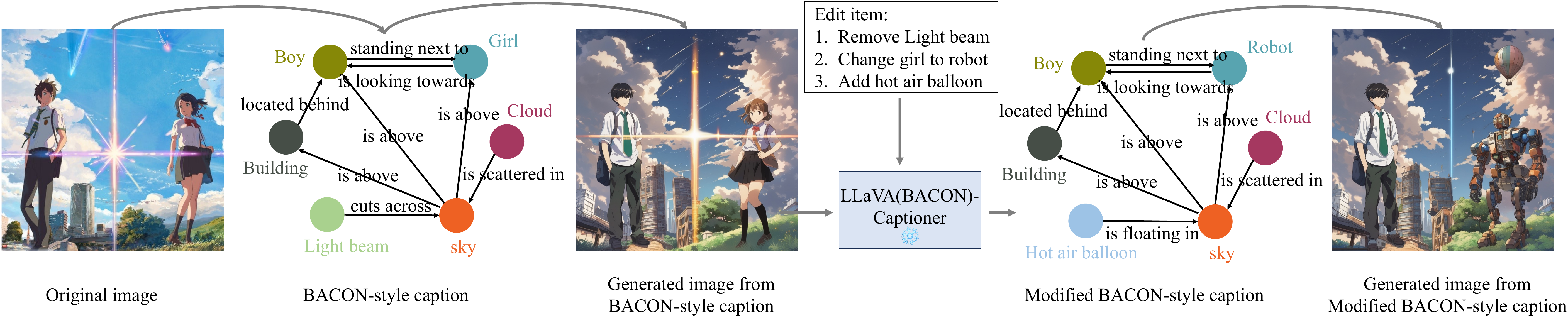}
\end{overpic}
\caption{
\textbf{An example of interactively modifying \methodabbr} using \captioner. 
}
\label{fig:edit}
\end{figure*}

\begin{figure*}[t]
\centering
\includegraphics[width=\textwidth]{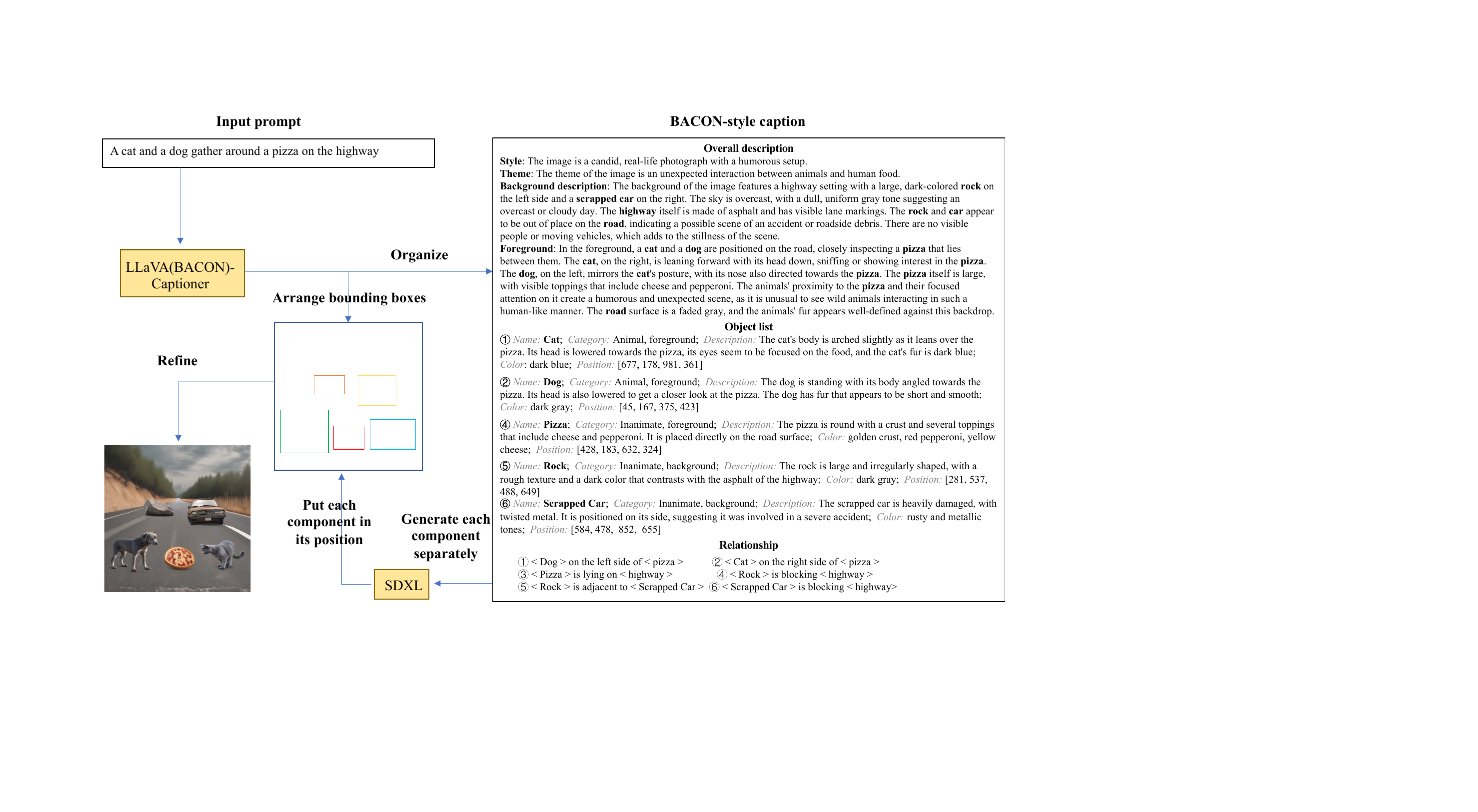}
\caption{
\textbf{An example of how \captioner boosts SDXL for image generation}. First, \captioner converts the prompt into the \methodabbr-style and arranges the bounding boxes of all objects. Then, each component is generated separately by SDXL and then placed in its arranged position. Finally, the combined image is refined to produce the final image.
}
\label{fig:recaption_1}
\end{figure*}

\begin{figure*}[t]
\centering
\includegraphics[width=\textwidth]{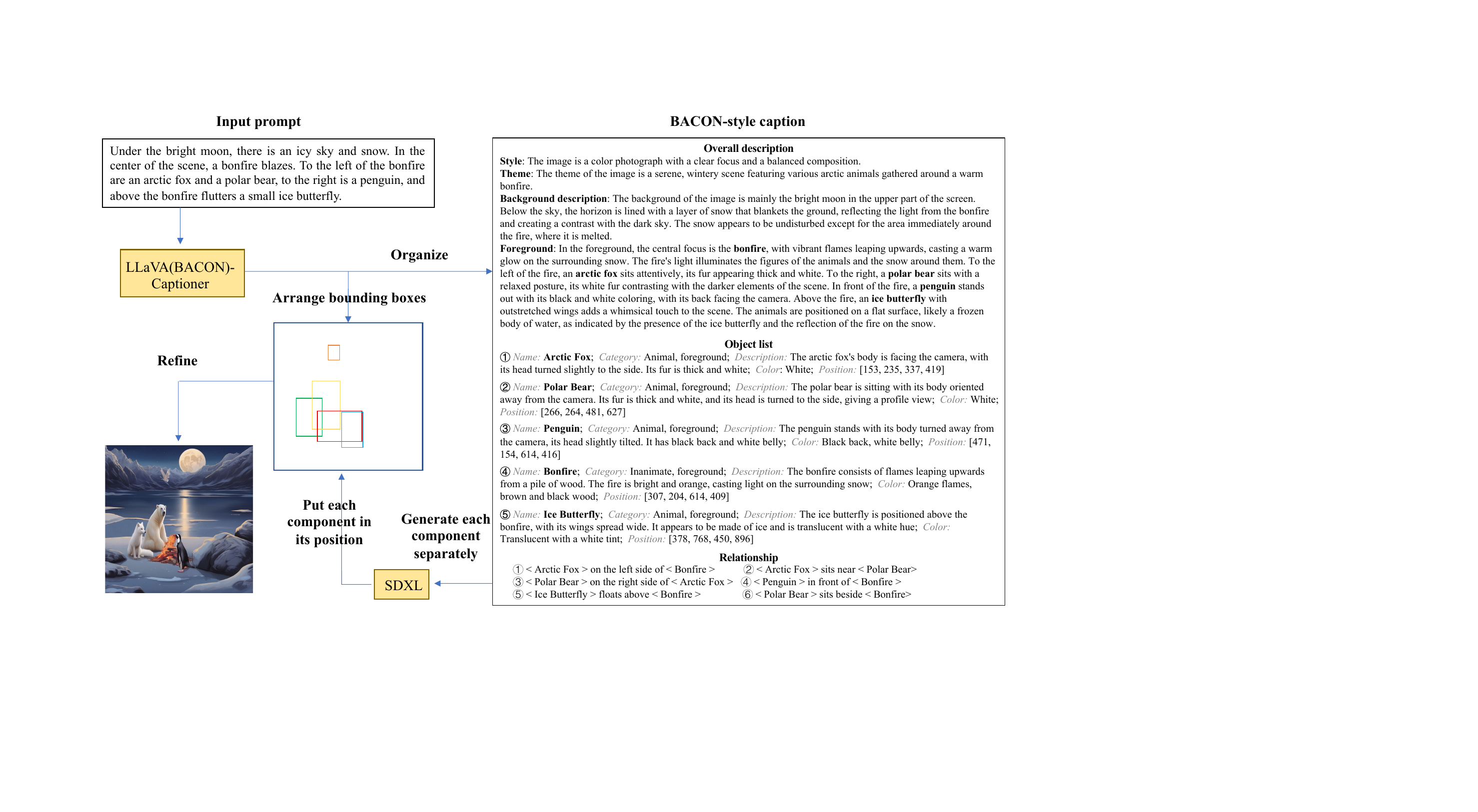}
\caption{
\textbf{An example of how \captioner boosts SDXL for image generation}. First, \captioner converts the prompt into the \methodabbr-style and arranges the bounding boxes of all objects. Then, each component is generated separately by SDXL and then placed in its arranged position. Finally, the combined image is refined to produce the final image.
}
\label{fig:recaption_2}
\end{figure*}

In addition to generating \methodabbr-style captions from images, the trained \captioner excels in several additional functions, including interactively editing \methodabbr-style captions by requesting desired changes from the \captioner, transforming ordinary prompts into \methodabbr-style captions, and planning the positions of objects within the object list.
First, as illustrated in \cref{fig:edit}, \textbf{the \captioner enables interactive editing of \methodabbr-style captions}, thereby influencing the image generation process.
Besides and remarkably, without requiring any fine-tuning, \textbf{the \captioner can convert a standard prompt into a \methodabbr-style caption}. 
This capability is particularly important for image generation, given the challenges of manually providing \methodabbr-style prompts.
Furthermore, \textbf{the \captioner can effectively arrange the positions of objects within the object list}. Examples of using \captioner to organize prompts and arrange the positions of objects for image generation can be found in \cref{fig:recaption_1,fig:recaption_2}. We quantitatively evaluate the planning capabilities of the \captioner against the leading LayoutGPT~\citep{feng2024layoutgpt} on the MSCOCO dataset~\citep{lin2014microsoft} and our \methodabbr datasets, employing mIoU, precision, and recall metrics~\citep{feng2024layoutgpt}.
The results presented in \cref{tab:plan} demonstrate that the \captioner outperforms LayoutGPT across both evaluated datasets.

\begin{table}[t]
\caption{
\textbf{Comparison of the VLM trained on QA data} derived from \dset (\methodabbr-13B, Ours) with other VLMs across 7 general benchmarks.
}
\vspace{-20pt}
\label{tab:vlm_understanding}
\begin{center}
\setlength{\tabcolsep}{1.8pt}
\footnotesize
\begin{tabular}{l|ccccccc}
\toprule
Model & GQA & SQA$^{I}$ & POPE & MMB & MMB$^{CN}$ & SEED & MM-Vet\\
\midrule
Qwen-VL                     & 59.3$^*$   & 67.1  & - & 38.2 & 7.4 & 56.3 & -  \\
Qwen-VL-Chat                & 57.5$^*$   & 68.2  & - & 60.6 & 56.7 & 58.2 & -  \\
LLaVA-13B                   & 63.3   & 71.6  & 85.9 & 67.7 & 63.6 & 61.6 & 35.4  \\
VILA-13B                    & 63.3   & 73.7  & 84.2 & 70.3 & 64.3 & 62.8 & 38.8  \\
\midrule
Ours      & \textbf{63.5}    & \textbf{91.3}  & \textbf{88.0} & \textbf{74.6} & \textbf{68.2} & \textbf{65.9}  & \textbf{41.6}  \\
\bottomrule
\end{tabular}
\vspace{-15pt}
\end{center}
\end{table}

\section{Additional details of experiments}\label{sec:app-experiments}

In this section, we provide more details about \cref{sec:exp}. This includes details of the five experiments discussed respectively in \cref{subsub:exp_ovsgg} (Analyzing Object \& Relation Accuracy using 
Open-vocabulary scene graph generation), \cref{subsubsec:exp_user_study} (Analyzing Precision \& recall using user study), \cref{subsubsec:exp_pointqa} (Zero-shot region-based question answering), \cref{subsubsec:exp_video_caption} (Multi-object video tracking and dense video captioning), and \cref{subsubsec:exp_image_generation} (Image generation). Each experiment is presented in its own subsection, covering aspects that are omitted from the main text, including detailed evaluation metrics, the methods employed by the models to utilize \methodabbr-style captions, and other relevant details.

\subsection{Details of evaluating \captioner on open-vocabulary scene-graph generation benchmark}\label{sec:app-exp-ovsgg}

As discussed in \cref{subsub:exp_ovsgg}, we use the open-vocabulary scene graph generation benchmark to evaluate the performance of \captioner. In this section, we introduce more details about the evaluation method from two perspectives, the dataset and the evaluation metrics.

\noindent\textbf{Visual Genome.} 
As introduced in the main paper, Visual Genome (VG)~\citep{krishna2017visual} is employed for evaluation. VG is an open-vocabulary dataset; however, most current scene graph generation (SGG) models typically consider only a limited number of categories. Consequently, researchers often treat it as a dataset with a restricted set of categories. Specifically, they usually identify the 70 or 150 most frequent noun classes, along with the 50 most common predicates, to create a filtered dataset. In our case, since we are working on an open-vocabulary scene graph generation (OV-SGG) task, we treat the VG dataset as an open-vocabulary resource and utilize only the triplets containing nouns or predicates that fall outside the commonly used set.

\noindent\textbf{Evaluation metrics.} 
Traditional SGG tasks often utilize recall-related metrics to evaluate performance, specifically assessing how many (subject-predicate-object) triplets present in an image are correctly predicted. Previous models typically perform classification tasks within a fixed set of categories and use the confidence of those classifications to obtain the top K predictions with the highest likelihood. Any ground truth triplets that appear within the top $K$ predictions are considered correctly predicted.

However, in an open-vocabulary setting, the possibilities are virtually infinite, making it challenging to identify the top K predictions. Therefore, we employ the CLIP~\citep{radford2021learning} score to determine whether a prediction is correct. To do this, we construct a string for each triplet in the format "$\{$subject$\}_\{$predicate$\}_\{$object$\}$" and calculate the CLIP similarity between the prediction and the ground truth. If we find that a ground truth has a CLIP similarity score with the prediction that exceeds a specified threshold (0.9 in this case), and the Intersection Over Union (IOU) of the subject and object positions between the prediction and the ground truth also surpasses another threshold (0.5 here), we consider it a correct prediction. Consequently, we can calculate the recall score.

\subsection{Additional details for user study}\label{sec:app-exp-user}
As discussed in \cref{subsubsec:exp_user_study}, calculating precision and recall involves identifying the objects predicted by different captioners. For captioners other than \captioner, this can be challenging, as directly extracting nouns may include many terms that cannot be considered objects. To address this issue, we utilize VLMs for the task. Specifically, we input the captions into the VLMs and request them to extract the objects contained within. In contrast, this process is straightforward for \captioner, as \methodabbr explicitly provides a list of objects. This emphasizes the superiority of \captioner.

\subsection{Additional details of \methodabbr-style captions assist LLaVA in zero-shot region-based question answering.}\label{sec:app-exp-regionqa}

\begin{figure*}[t]
  \centering
  \begin{overpic}[width=\textwidth]{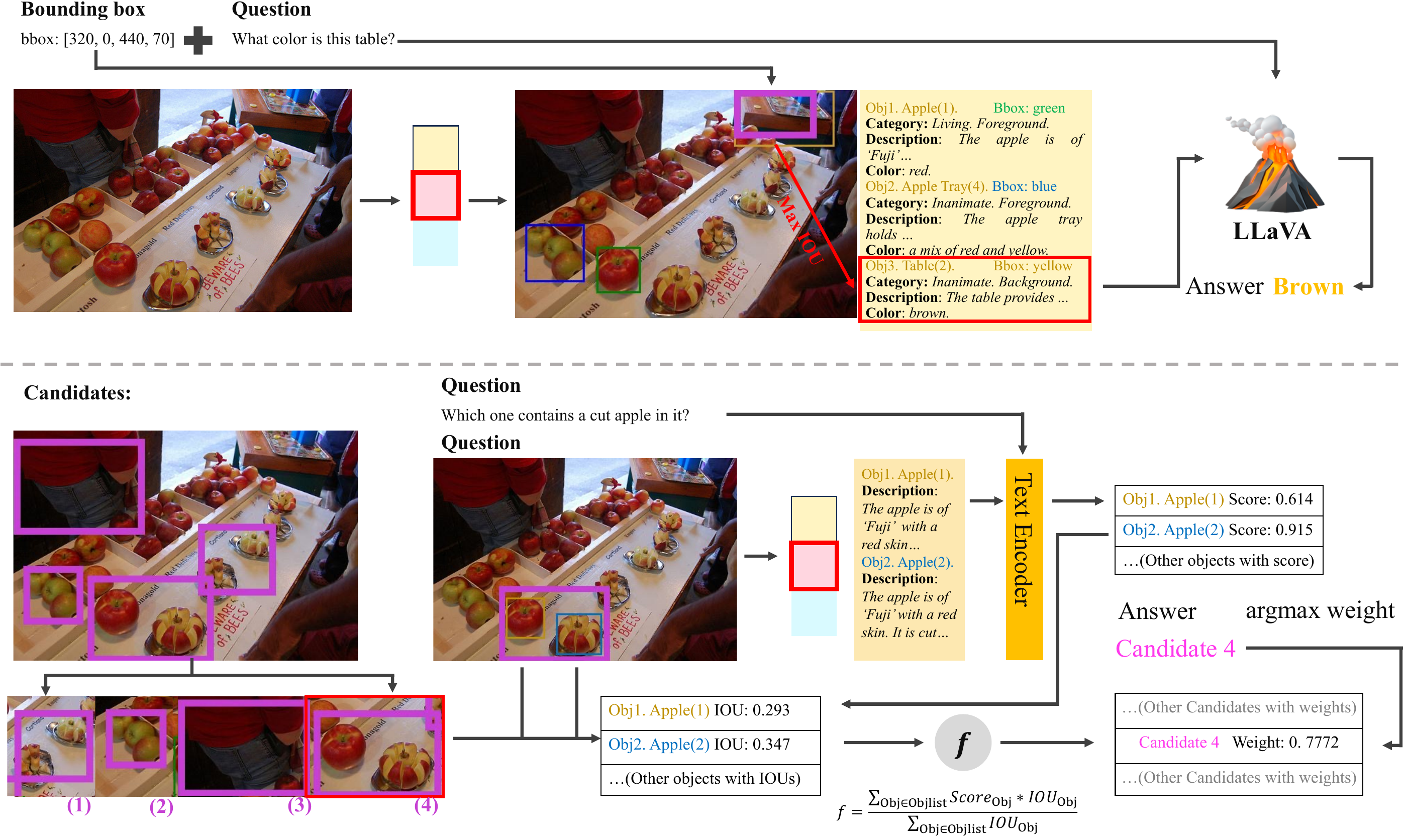}
    \put(40,35){\textcolor{black}{\small\text{(a) Zero-shot PointQA task}}}
    \put(40, -1){\textcolor{black}{\small\text{(b) Zero-shot PointingQA task}}}
  \end{overpic}
  \caption{
    \textbf{An illustrative diagram depicting how \methodabbr-style captions aid downstream models} in executing zero-shot PointQA and zero-shot PointingQA tasks. In (a) the zero-shot PointQA task, the description of the object that has a significant overlap with the target region is used to characterize that region. This regional description is then input into a QA model to answer questions related to the area. In (b) the zero-shot PointingQA task, object descriptions provided by \methodabbr are used to calculate similarity scores with the input question, generating scores for each object. Based on the overlap between object positions and candidate regions, a weighted sum of all object scores is computed to assign scores to candidate regions; the region with the highest score is then selected as the prediction.
  }
  \label{fig:region_based_qa}
\end{figure*}

\begin{figure*}[t]
\centering
\includegraphics[width=\textwidth]{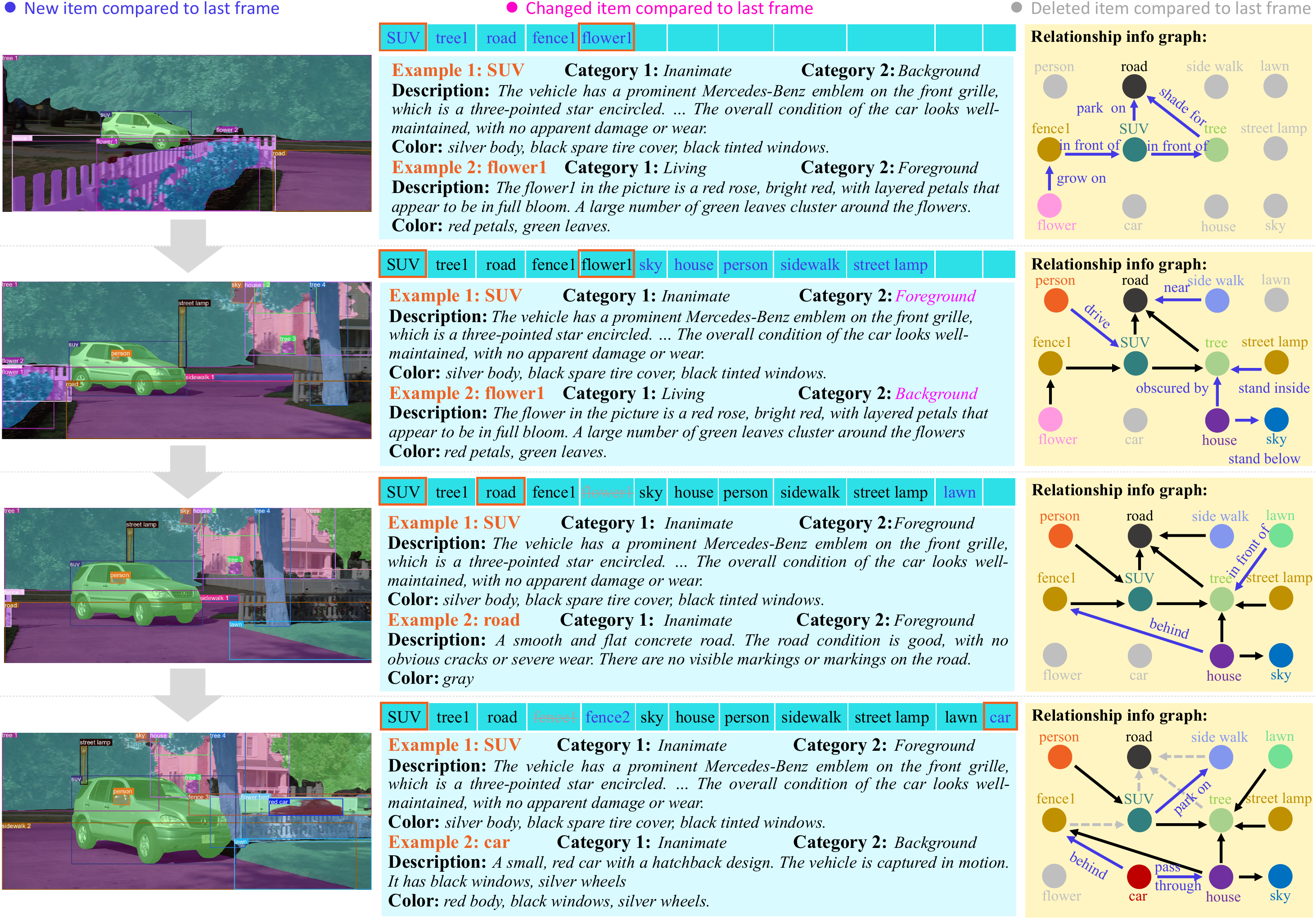}
\caption{
\textbf{An example of \captioner assists SAM-2 on dense video captioning}.
}
\label{fig:app_video_caption_1}
\end{figure*}

\begin{figure*}[t]
\centering
\includegraphics[width=\textwidth]{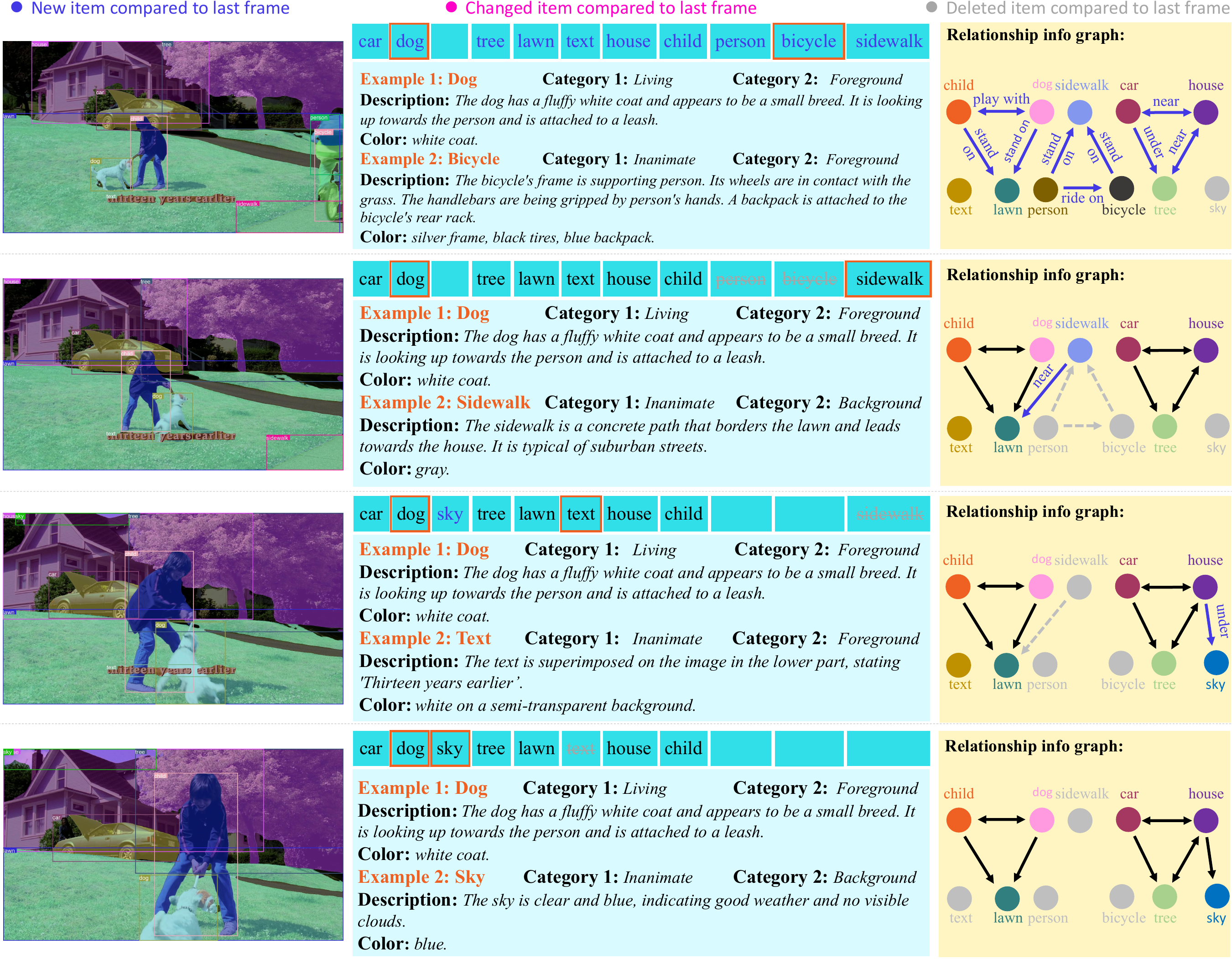}
\caption{
\textbf{An example of \captioner assists SAM-2 on dense video captioning}.
}
\label{fig:app_video_caption_2}
\end{figure*}

LLaVA lacks the capability to connect information to its corresponding regions without fine-tuning, making it challenging for LLaVA to perform region-based question-answering tasks. Fortunately, \methodabbr-style captions provide bounding boxes for all objects along with their descriptions, enabling the connection between descriptions and regions.
In \cref{subsubsec:exp_pointqa}, we have introduced the benchmarks used, the evaluation metrics employed, and provided the experimental results. Here, we present the detailed method by which LLaVA utilizes \methodabbr-style captions to perform zero-shot region-based question answering, including both zero-shot PointQA and zero-shot PointingQA.

\noindent\textbf{Zero-shot PointQA}
Zero-shot PointQA is designed to answer questions related to specific regions of an image solely based on the image caption instead of the image itself. Given a target area, we can create a description relevant to that location by combining the descriptions of different objects based on their positions. Specifically, as illustrated in \cref{fig:region_based_qa}(a), we calculate the Intersection Over Union (IOU) between the target area and the positions of all objects. By combining the descriptions of objects with large overlaps, we generate a description that is closely linked to the target area. Subsequently, we input this region-based description into the question-answering model to derive the answer to the question.

\noindent\textbf{Zero-shot PointingQA}
The Zero-shot PointingQA task involves selecting the most appropriate region from a set of candidate areas based on a given textual request and the image's description, rather than the image itself. As illustrated in \cref{fig:region_based_qa}(b), the method consists of three steps: 1) First, we compute the CLIP similarity between each object's description and the input textual prompt, resulting in scores for each object. The more relevant an object is to the text description, the higher its score will be. 2) Next, we calculate scores for each candidate region by weighting the sum of object scores based on the overlap between the candidate region and the object's location. The greater the overlap with the candidate area, the larger the contribution of that object's score. 3) In the final step, we select the region with the highest score as the answer.

\subsection{Additional details of \methodabbr-style captions assist SAM-2 in dense video captioning.}\label{sec:app-exp-videotrack}

\noindent\textbf{Methods.} 
As discussed in Section 4.2.3, \methodabbr-style captions can assist SAM-2 in performing dense video captioning tasks by providing bounding boxes as indicators and supplying the descriptions that SAM-2 lacks.
While this method is a reliable approach for video captioning, it still encounters two challenges. The first challenge involves managing newly appeared objects in the video, and the second is how to ensure that the caption content evolves as the video progresses. To address these challenges: 1) We begin by uniformly sampling several frames from the video and employing tracking from each selected frame to monitor the entire video. We then consolidate the tracking results across different frames and assign the same ID to objects with exceptionally high mask overlap. 2) Additionally, we utilize T5~\citep{raffel2020exploring} as the text encoder to compare descriptions of the same object or scene segment across frames. Portions of the text with high similarity scores are deemed stable, while those with low similarity scores are identified as having changed.

\noindent\textbf{More results.} 
We provide more examples in \cref{{fig:app_video_caption_1,fig:app_video_caption_2}}

\subsection{Additional details of \methodabbr-style captions assist SDXL in image generation.}\label{sec:app-exp-image-generation}

\noindent\textbf{Methods.} 
Even as one of the most renowned models for text-to-image generation, SDXL often struggles to understand complex prompts and generate precise images accurately. This is primarily because SDXL employs CLIP for text understanding, which limits its ability to comprehend the text. Fortunately, by breaking down complex texts into basic elements, \methodabbr-style captions can significantly assist SDXL in simplifying complex tasks. The specific method by which \methodabbr-style captions enhance SDXL for image generation can be divided into three steps:
\begin{itemize}
    \item \textbf{Step 1}: Given a natural prompt for generation, the trained \captioner converts it into a \methodabbr-style caption and arranges the bounding boxes for all objects listed in the caption.
    \item \textbf{Step 2}: SDXL separately generates all components, including the background and each object. To create the background, SDXL uses the background description. Besides, it relies on the detailed descriptions of objects to generate them.
    \item \textbf{Step 3}: For the generated image of the objects, we extract the main components by segmenting them from the image using SAM~\citep{kirillov2023segment}. These components are then combined according to their arranged positions from Step 2. Finally, the combined image is refined using commonly employed refinement methods, including Anydoor~\citep{chen2024anydoor}, Collage Diffusion~\citep{sarukkai2024collage}, inpainting~\citep{rombach2022high}, and SDEdit~\citep{meng2021sdedit}. Some methods are applied during the combination process, while others are utilized afterward.
\end{itemize}

\begin{figure*}[t]
\centering
\includegraphics[width=\textwidth]{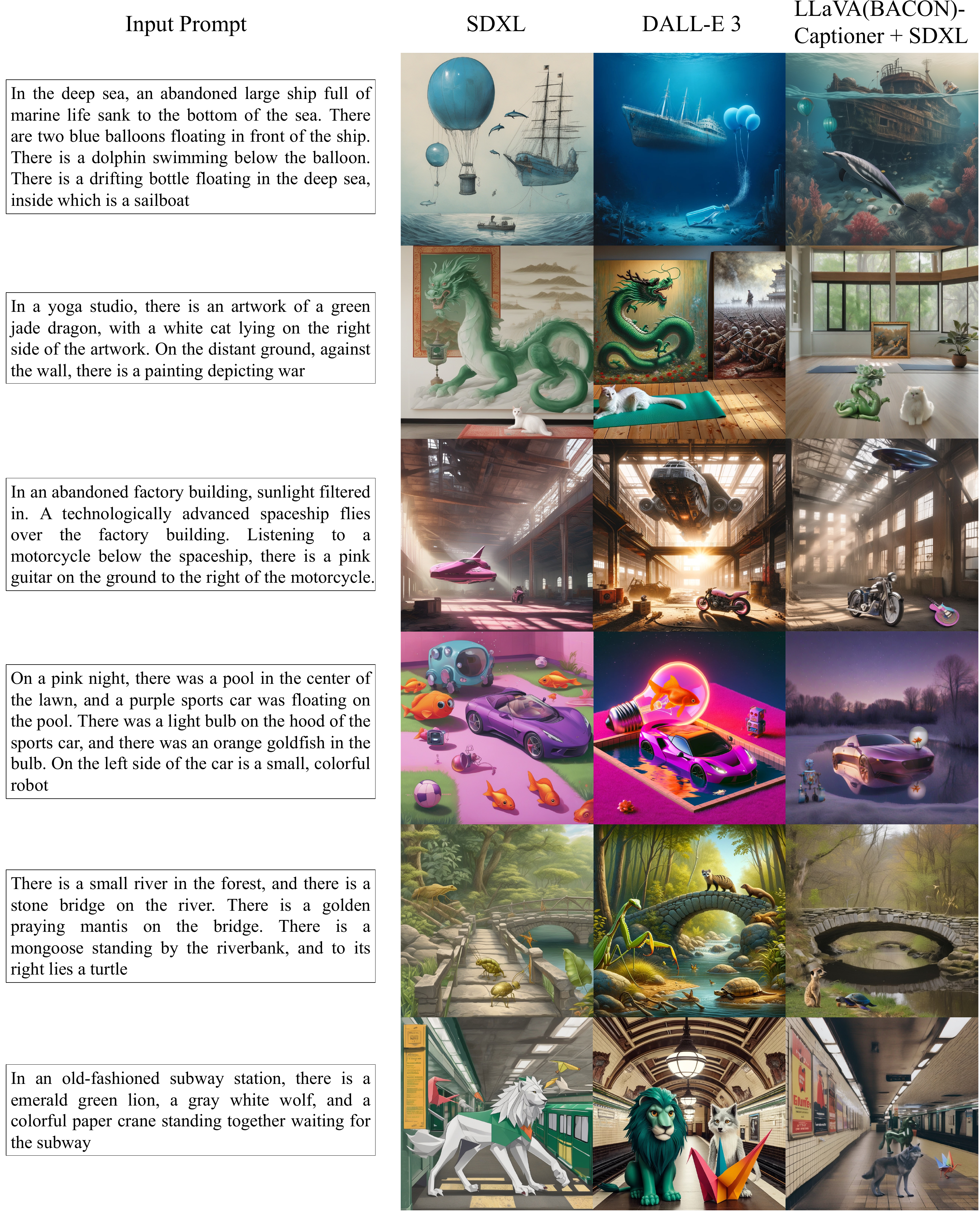}
\caption{
\textbf{Additional examples of \methodabbr boosting SDXL on image generation.}
}
\label{fig:app_image}
\end{figure*}

\noindent\textbf{More results.}
We provide more examples in \cref{fig:app_image}

\subsection{\methodabbr-style captions enhance the multi-modal understanding capabilities of VLMs.}\label{sec:app-exp-multimodal-understanding}

In this section, we show that the multi-modal understanding capabilities of VLMs can be further improved by training on \methodabbr-style captions. To demonstrate this, we fine-tune an LLaVA-13B using \dset by utilizing QA data derived from them.
Specifically, the structured nature of the captions allows us to automatically generate multiple QA pairs from a single caption, such as 'Is object A to the right of object B?', 'Is object A red?', and 'Please describe object A.' 
We present comparison results of this VLM against previous models on commonly used benchmarks in \cref{tab:vlm_understanding}. Results suggest that data collected through BACON enhances the multi-modal understanding capabilities of VLMs.

\end{document}